\documentclass[11pt]{article}
\usepackage[]{ACL2023} 
\usepackage{times}
\usepackage{blindtext}
\usepackage{latexsym}
\usepackage{multirow}
\usepackage{multicol}
\usepackage{amsmath}
\usepackage{booktabs}
\usepackage{xcolor}
\usepackage{float}
\usepackage{algpseudocode}
\usepackage{listings}
\usepackage{color}
\usepackage{amssymb}
\usepackage{footnote}
\usepackage{supertabular}
\usepackage{titlesec}
\usepackage[toc,page,header]{appendix}

\DeclareCaptionLabelFormat{andtable}{#1~#2  \&  \tablename~\thetable}
\usepackage{graphicx}
\usepackage[flushleft]{threeparttable}
\definecolor{backcolour}{rgb}{0.95, 0.95, 0.96}
\lstdefinestyle{mystyle}{
    backgroundcolor=\color{backcolour},   
    showtabs=false,                  
    tabsize=2
}
\usepackage{enumerate}
\usepackage{enumitem}
\newlist{compactitem}{itemize}{3}
\setlist[compactitem]{topsep=0pt,partopsep=0pt,itemsep=0pt,parsep=0pt,leftmargin=\parindent}
\setlist[compactitem,1]{label=\textbullet}
\setlist[compactitem,2]{label=---}
\setlist[compactitem,3]{label=*}

\newlist{compactdesc}{description}{3}
\setlist[compactdesc]{topsep=0pt,partopsep=0pt,itemsep=0pt,parsep=0pt}

\newlist{compactenum}{enumerate}{3}
\setlist[compactenum]{topsep=0pt,partopsep=0pt,itemsep=0pt,parsep=0pt}
\setlist[compactenum,1]{label=\arabic*.}
\setlist[compactenum,2]{label=\alph*.}
\setlist[compactenum,3]{label=\roman*.}

\usepackage{longtable}
\usepackage[T1]{fontenc}
\usepackage[utf8]{inputenc}
\usepackage{microtype}
\usepackage{inconsolata}
\usepackage[capitalize,nameinlink]{cleveref}
\usepackage{subcaption}
\makeatletter
\renewcommand{\sectionautorefname}{\S\@gobble}
\renewcommand{\sectionautorefname}{\S\@gobble}
\renewcommand{\subsectionautorefname}{\S\@gobble}
\renewcommand{\sectionautorefname}{\S\@gobble}
\renewcommand{\subsectionautorefname}{\S\@gobble}
\makeatother

\usepackage{graphicx}
\usepackage{tikz}
\usepackage{forest}
\usepackage{tikz-qtree}
\usetikzlibrary{trees,positioning,shapes,shadows,arrows.meta}

\newcommand{\tabitem}{~~\llap{\textbullet}~~}

\newcommand{\ensuretext}[1]{#1}
\newcommand{\nertcomment}[4]{\ensuretext{\textcolor{#3}{[\ensuretext{\textcolor{#3}{\ensuremath{^{\textsc{#1}}_{\textsc{#2}}}}} #4]}}}

\newcommand{\DC}[1]{\nertcomment{D}{C}{blue}{#1}}

\newcommand{\macro}[1]{\textcolor{black}{#1}} 

\newcommand\benchmark{\macro{\textsc{DialectBench}}}

\newcommand{\dialect}{\macro{variety}}
\newcommand{\dialects}{\macro{varieties}}
\newcommand{\Dialect}{\macro{Variety}}
\newcommand{\Dialects}{\macro{Varieties}}
\newcommand{\repr}{\macro{representative}}
\newcommand{\Repr}{\macro{Representative}}
\newcommand{\dial}{\macro{var.}}
\newcommand{\cluster}{\macro{cluster}}

\newcommand{\clusters}{\macro{clusters}}
\newcommand{\Cluster}{\macro{Cluster}}
\newcommand{\Lingu}{\macro{\textit{Linguistic utility}}}
\newcommand{\lingu}{\macro{\textit{linguistic utility}}}
\newcommand{\Demu}{\macro{\textit{Demographic utility}}}
\newcommand{\demu}{\macro{\textit{demographic utility}}}
\newcommand{\Clusters}{\macro{Clusters}}
\newcommand{\clus}{\macro{cl.}}
\newcommand{\gap}{\macro{\ensuremath{\mathcal{G}}}}
\newcommand{\gapiii}[3]{\macro{\ensuremath{\mathcal{G}_{#1}( #2, #3)}}}
\newcommand{\score}[2]{\macro{\ensuremath{\mathcal{S}_{#1}(#2)}}}
\newcommand{\std}[1]{\macro{\bar{#1}}}

\title{\benchmark: \\[0.3ex] A NLP Benchmark for Dialects, Varieties, and Closely-Related Languages}

\author{Fahim Faisal\textsuperscript{$\alpha$,}\thanks{Equal contribution.} \quad Orevaoghene Ahia\textsuperscript{$\beta$,*} \quad Aarohi Srivastava\textsuperscript{$\gamma$} \quad Kabir Ahuja\textsuperscript{$\beta$} \\
\textbf{David Chiang}\textsuperscript{$\gamma$} \quad 
\textbf{Yulia Tsvetkov}\textsuperscript{$\beta$} \quad 
\textbf{Antonios Anastasopoulos}\textsuperscript{$\alpha,\delta$}\\[1ex]
\textsuperscript{$\alpha$}George Mason University\quad
\textsuperscript{$\beta$}University of Washington\quad
\textsuperscript{$\gamma$}University of Notre Dame\\
\textsuperscript{$\delta$}Archimedes Research Unit, RC Athena, Greece\\[1ex]
\texttt{\{ffaisal,antonis\}@gmu.edu} \quad \texttt{\{oahia,kahuja,yuliats\}@cs.washington.edu} \\ \quad \texttt{\{asrivas2,dchiang\}@nd.edu}
  }

\begin{document}
\maketitle
\begin{abstract}
Language technologies should be judged on their usefulness in real-world use cases. An often overlooked aspect in natural language processing (NLP) research and evaluation is language variation in the form of non-standard dialects or language varieties (hereafter, \dialects). Most NLP benchmarks are limited to standard language varieties. To fill this gap, we propose \benchmark{}, the first-ever large-scale benchmark for NLP on \dialects, which aggregates an extensive set of task-varied \dialect{} datasets (10 text-level tasks covering 281 \dialects). This allows for a comprehensive evaluation of NLP system performance on different language \dialects. We provide substantial evidence of performance disparities between standard and non-standard language varieties, and we also identify language \clusters{} with larger performance divergence across tasks.
We believe \benchmark{} provides a comprehensive view of the current state of NLP for language \dialects{} and one step towards advancing it further. \footnote{More information can be found at the following:\\\begin{tabular}[t]{l@{}}Code/data: \scalebox{0.93}{\url{https://github.com/ffaisal93/DialectBench}} \\ Website: \scalebox{0.93}{\url{https://fahimfaisal.info/DialectBench.io}}\end{tabular}}

\end{abstract}

\section{Introduction}
Benchmarking is important for tracking the progress the field of natural language processing (NLP) has made in various tasks. In the past few years, large-scale multilingual benchmarks like XTREME~\cite{pmlr-v119-hu20b}, XTREME-R~\cite{ruder-etal-2021-xtreme}, and XGLUE~\cite{liang-etal-2020-xglue-fixed} have played a pivotal role in evaluating the multilingual capabilities of NLP models. These efforts have sought to make model evaluation more accessible to researchers and representative of a variety of languages \cite{song2023globalbench}. However, most benchmarks have focused on the standard varieties of languages, largely neglecting non-standard dialects and language varieties \cite{blasi-etal-2022-systematic}.

\begin{figure}[!t]
    \centering
    \includegraphics[width=.4\textwidth]{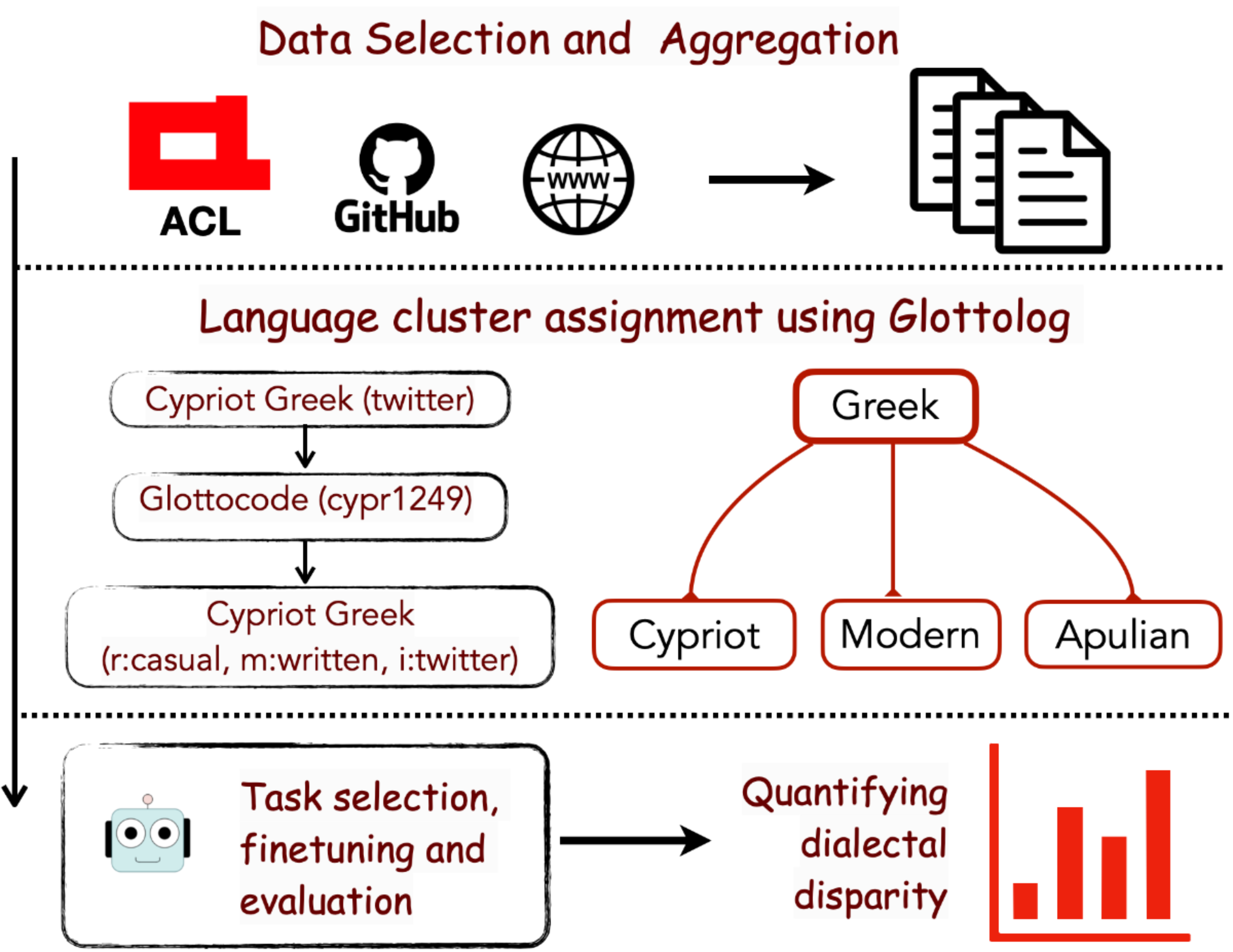}
    \caption{\benchmark{} Evaluation Suite.}
    \label{fig:framework}
    \vspace{-1em}
\end{figure}

We refer to non-standard dialects and language varieties simply as \emph{\dialects}, and sometimes include low-resource related languages, writing system variants, and other kinds of variation. 
\Dialects{} contain subtle but notable variations in vocabulary, pronunciation, orthography and grammar, reflecting regional, social, and cultural differences \cite{chambers1998dialectology}. The non-standard nature of these language varieties oftentimes contributes to the scarcity of substantial datasets that accurately capture these variations \cite{hedderich-etal-2021-survey}. As a result, they have often been absent from widely adopted benchmarks, even from admirable efforts like XTREME-up~\cite{ruder2023xtremeup}, GLUECoS~\cite{khanuja-etal-2020-gluecos} and CreoleVal ~\cite{lent2023creoleval}, which focus on under-resourced, code-switched, and creole languages, respectively.
\begin{figure}[!t]
    \centering
    \includegraphics[width=.5\textwidth]{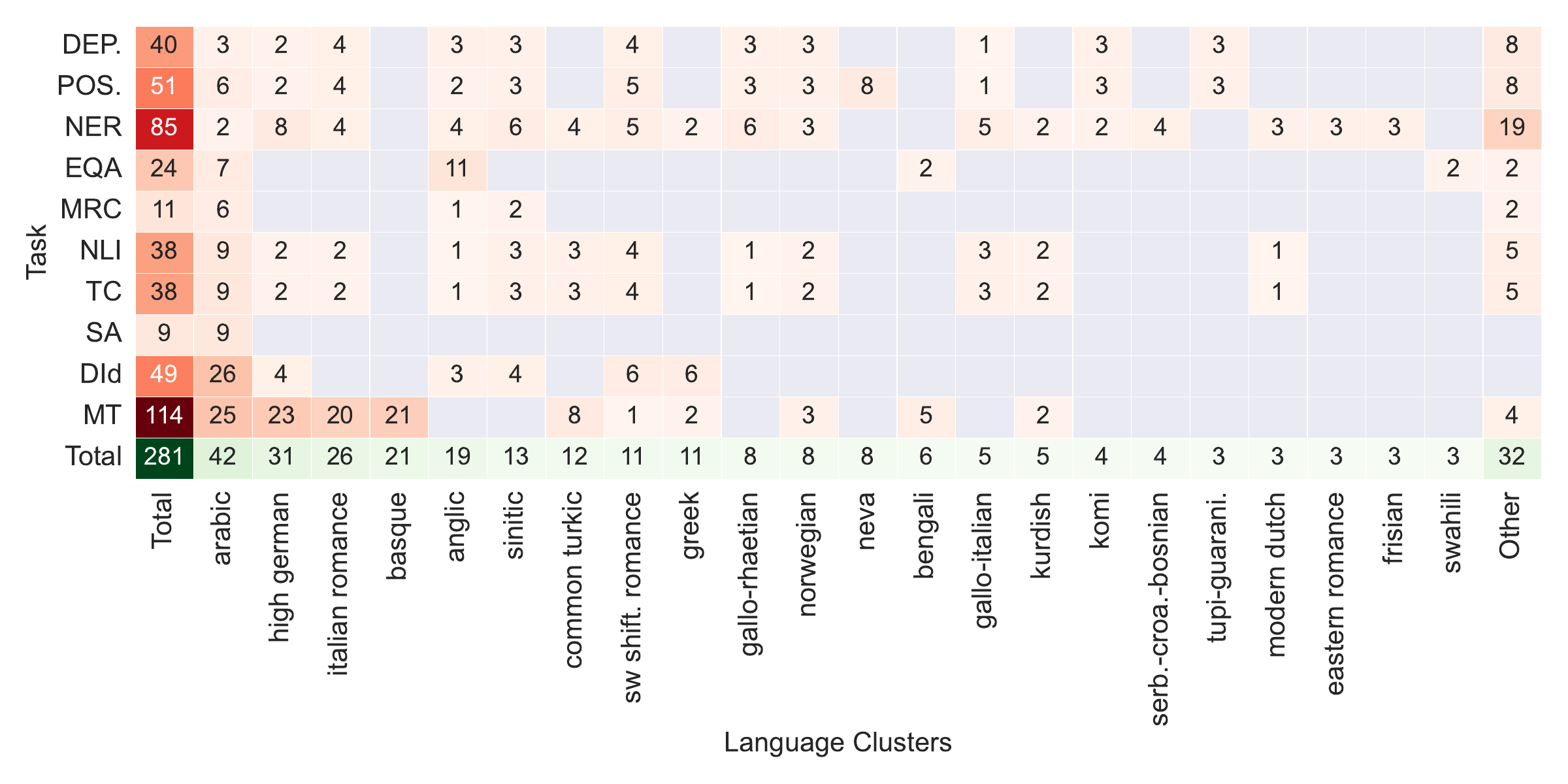}
    \vspace{-2em}
    \caption{\benchmark{} language \clusters{} with their \dialect{} counts per task. "Other" encompasses 18 clusters (full \cluster{} list in Appendix~\cref{lang_list}).}
    \label{fig:dial_count}
\end{figure}
It is currently challenging to accurately test the robustness of multilingual models on a large suite of \dialects{} without establishing an NLP evaluation framework covering multiple language \clusters{} (each of which contains standard languages alongside its closely related \dialects). 

To this end, we create \benchmark, a large-scale benchmark covering 40 language \clusters{} with 281 \dialects, spanning 10 NLP tasks. We observe that the performance disparity between different \dialects{} of the same language \cluster{} becomes more pronounced when we shift from zero-shot evaluation to fine-tuning on \dialect{} data, because of uneven data availability across \dialects. 
Certain language \clusters{} exhibit varying performance across downstream tasks within the same category, due to low-resource limitations. 
Additionally, we improve the dialectal task coverage for natural language inference by constructing a \textit{translate-test} evaluation dataset. Putting these all together, \benchmark{} serves as a comprehensive suite that attains a multifaceted purpose: identifying broader limitations in dialectal NLP, while reflecting on potential areas for improvement.

\section{\benchmark}
\begin{table*}[!t]
    \centering
    \tiny
    \begin{tabular}{@{}p{1.2cm}|p{1.5cm}@{}p{.8cm}p{11cm}@{}}
    \toprule
    Category & Task & Metric & Source Dataset \\
    \midrule
     Structured &DEP parsing&UAS&Universal Dependency~\cite{11234/1-4611}, TwitterAAE~\cite{blodgett-etal-2018-twitter}, Singlish~\cite{wang-etal-2017-universal-fixed}\\
     Prediction&POS tagging&F1&Universal Dependency~\cite{11234/1-4611}, Singlish~\cite{wang-etal-2017-universal-fixed}, Noisy Dialects~\cite{blaschke-etal-2023-manipulating}\\
     &NER & F1 & Wikiann~\cite{pan-etal-2017-cross,rahimi-etal-2019-massively}, Norwegian NER~\cite{johansen2019ner} \\
     \midrule
     Classification& DId & F1 & MADAR~\cite{bouamor-etal-2018-madar}, DMT~\cite{jauhiainen-etal-2019-discriminating}, Greek~\cite{8554709}, DSL-TL~\cite{zampieri2023language}, Swiss Germans~\cite{articleswiss}\\
     &SA&F1&TSAC~\cite{medhaffar-etal-2017-sentiment},
TUNIZI~\cite{fourati-etal-2021-introducing},
DzSentiA~\cite{9068897},
SaudiBank~\cite{alqahtani-etal-2022-customer},
 MAC~\cite{Garouani_MAC},
 ASTD~\cite{nabil-etal-2015-astd},
AJGT~\cite{10.1007/978-3-319-60042-0_66},
OCLAR~\cite{8716394} \\
&TC & F1 & SIB-200~\cite{adelani2023sib200}\\
&NLI& F1 & XNLI~\cite{conneau-etal-2018-xnli} translate-test\\
\midrule
Question & MRC& F1 & Belebele~\cite{bandarkar2023belebele}\\
Answering& EQA& Span F1 &SDQA~\cite{faisal-etal-2021-sd-qa}\\
\midrule
Generation & MT& BLEU&CODET~\cite{alam2023codet}, TIL-MT~\cite{mirzakhalov2021turkic}\\

  \bottomrule
    \end{tabular}
    \caption{The tasks and data sources of \benchmark{} (Detailed discussion:  \cref{{sec:tasks_benchmark}}).}
    \label{tab:task}
\end{table*}
\benchmark{} is a benchmark created to unify dialectal datasets across different tasks to foster research on language varieties and non-standard dialects. Below we describe the design choices we undertook to achieve this goal. This includes our language \dialect{} and task selection procedures, data collation methods, and evaluation principles. 

\paragraph{Variety Selection} 
We first looked through papers published in the ACL Anthology\footnote{\url{https://aclanthology.org}} from the last 10 years to find usable language resources, as well as commonly used online data repositories~\cite{githubGitHubRichardLittlowresourcelanguages}. We selected languages that have well-established, high-resourced varieties. Varieties may vary by location, ethnicity, or other factors. We also found instances where \dialects{} are classified by writing system or even by genre (e.g., Twitter). When varying by location, \dialects{} may be classified by different datasets at different levels of \emph{granularity}, sometimes country, region, or city. In some cases, we found resources with two or more \dialects{} within one dataset (e.g., the \texttt{UD\_Portuguese-Bosque} dependency treebank~\cite{depling-2017} includes examples from both European and Brazilian Portuguese variants. To incorporate all these cases under one paradigm, we formulate a \cluster-\dialect{} mapping procedure.

\paragraph{\Cluster-\Dialect{} Mapping} We construct several language \clusters{} comprising of both high-resourced \dialects{} and their low-resourced counterparts. We use the Glottolog language database~\cite{ldl-glottolog} to define \clusters{} and assign varieties as outlined in \autoref{fig:framework}. This design choice enables us to keep \dialects{} that are closely related in terms of either mutual intelligibility, phylogenetic similarity or geographic proximity within the same \cluster{}. Hence, all \cluster{} \dialects{}  always root back to the closest common linguistic ancestor and the whole \cluster{} maps to an established phylogenetic subtree. For example, Fiji Hindi and Hindi, with Hindustani\footnote{\url{https://glottolog.org/resource/languoid/id/hind1270}} as their closest common ancestor, are placed in the Hindustani~\cluster{}.

We primarily use the Glottocode language identification scheme~\cite{swj-glottocodes}, 
ensuring a standardized naming scheme across all \dialects{}. 
For instance, AAE \dialect{} from TwitterAAE~\cite{blodgett-etal-2018-twitter} dependency parsing dataset is renamed as \texttt{African American Vernacular English} with a corresponding Glottocode \texttt{afri1276}. In cases where Glottocodes are unavailable, like for spoken English from South India, we substitute with the immediate ancestor Glottocode (\texttt{indi1255}) and further categorize the \dialects{} using the following metadata identifiers:
\begin{compactenum}
\item Area (a): the region where the \dialect{} is spoken or where its dataset was collected.
\item Language register (r): frozen, formal, consultative, casual, and intimate.
\item Language mode (m): written, spoken, and signed language.
\item Orthography (o): In \benchmark{} this is only specific to Sinitic \dialects{}. This could be either traditional or simplified.
\item Identifier (i): Dataset-specific metadata, could be domain (eg. twitter).
\end{compactenum}

We encapsulate all this information, into a naming convention, and use the template: \{Glottocode name\}-(a:\{\},r:\{\},m:\{\},o:\{\},i:\{\}). \footnote{For example, \textit{mandarin chinese (a:mainland, o:simplified} refers to Mandarin Chinese (mand1415) spoken in Mainland China and written in simplified characters.}

\paragraph{\Cluster{} \Repr} Each \cluster{} will often have a high-resourced \dialect{} usually with the largest speaker population. We choose this high-resourced \dialect{} as the \textit{\cluster{} representative}. This selection might vary across downstream tasks depending on the data availability. We primarily utilize this \repr{} \dialect{} to evaluate the performance gap across \cluster{} \dialects{}, and also rely on it for transfer-learning in resource-scarce settings. Sometimes, the members of a \cluster{} are considered closely related languages, and sometimes dialects; to avoid making this distinction, we refer to all the members of a \cluster{} simply as \emph{\dialects} of the \cluster{} \repr{}.

\paragraph{Task and Dataset Selection}
In selecting tasks, we maintain a balanced approach, promoting task diversity while also including tasks that require diverse levels of textual understanding. 
In the end, our complete list of tasks are as follows:
\begin{compactenum}
    \item Dependency parsing (DEP parsing)
    \item Parts-of-speech tagging (POS tagging)
    \item Named entity recognition (NER)
    \item Dialect identification (DId) 
    \item Sentiment analysis (SA)
    \item Topic classification (TC) 
    \item Natural language inference (NLI)
    \item Multiple-choice machine reading comprehension (MRC)
    \item Extractive question answering (EQA)
    \item Machine translation (MT)
\end{compactenum}

\noindent In \cref{tab:task}, we present the task and dataset details. We mostly keep the datasets in their originally published form (except for \dialects{} renaming). For NLI, we use the existing English test set of XNLI~\cite{conneau-etal-2018-xnli} and construct a multilingual dialect-focused translated evaluation dataset. We refer to this as translate-test NLI.

\paragraph{Evaluation Principles} 
On the ground level, we evaluate existing NLP systems on text-based tasks using standard evaluation metrics (e.g., UAS for parsing, F1 for classification tasks, BLEU for translation). 
At a global level, we believe a sustainable NLP system should be user-focused while providing substantial (i) \lingu{} and (ii) \demu{}~\cite{song2023globalbench, blasi-etal-2022-systematic}. \citet{blasi-etal-2022-systematic} defined the utility of a task and language, as the corresponding performance normalized by the best possible performance (usually human-level performance). \Demu{} considers the demand for a language technology within a specific language, where the demand is proportional to the number of speakers of that language. \Lingu{}, on the other hand, asserts that ``all languages are created equal'' regardless of the number of speakers, and hence all languages in the world should receive identical weights. 

Overall, we want to capture the performance gap between language \clusters{} (e.g., Anglic vs.~Italian Romance) as well as within language \clusters{} (e.g., Norwegian Bokmål vs.~Nynorsk). To attain this, we define the performance gap metrics in \autoref{subsec:gap_metrics} . We also vary the experimental settings in \autoref{subsec:experimental_setup}, including zero-shot and few-shot cross-lingual transfer, as well as fine-tuning with similar high-resource languages. This is essential given that we lack clean annotated data in many \dialects{}.

\section{Experiments}
\label{sec:experiments}

Here, we report the entire process involved in creating and evaluating baselines for all of the tasks and \dialects{} in \benchmark. Additionally, we define a \emph{dialectal gap} metric to analyze performance disparities within and across \clusters{}.

\subsection{Models}

We evaluate using two multilingual models: mBERT  \cite{devlin-etal-2019-bert} and XLM-R \cite{conneau-etal-2020-unsupervised} for all tasks except MT. For MT, we do zero-shot evaluation with NLLB \cite{nllbteam2022language}, using both the 600M and 1.3B variants. In addition, we use Mistral 7B~\cite{jiang2023mistral} to evaluate the current capability of LLMs on multilingual and dialectal understanding tasks. 
Our main goal is collating dialectal data across different languages and tasks under a single platform, hence we do not optimize for the best model performance. Rather, we focus on understanding and reporting the current state of performance on all \benchmark{} \dialects{}. 

\subsection{Training and Evaluation}\label{subsec:experimental_setup}
Training and evaluation procedures are largely determined by the availability of training or evaluation data for each task.


For any \cluster{} $C$, let $\std{C}$ be the highest-resourced \dialect{} (which is usually the \cluster{} \repr) of~$C$.
In addition, for any \dialect{} $v \in C$,  we write $\std{v}$ for the highest-resourced variety, that is, $\std{v} = \std{C}$. 
For any \dialects{}~$t$ and $v$, let $\score{t}{v}$ be the raw evaluation score of a system fine-tuned on $t$ and tested on $v$ (higher is better).

We use five 
general approaches for task-specific model training: 
\begin{compactenum}
\item \textbf{In-\dialect{} fine-tuning}\label{sec:in_lang_finetune} ($\score{v}{v}$): In cases where there is available training data for a \dialect{} $v$, we fine-tune the base model on $v$. This primarily applies to tasks such as POS tagging and dependency parsing. 

\item \textbf{In-\cluster{} fine-tuning} ($\score{\std{v}}{v}$): In-\dialect{} fine-tuning is 
quite resource-intensive when we have a large number of \dialects{} within a \cluster{} $C$. In such cases, we fine-tune the base model on $\std{C}$. Then we evaluate this model on each \dialect{} $v \in C$. This is the most common setting in our experiments, as it allows us to evaluate the dialectal performance gap without increasing the computation cost. For the DId task, we typically use a dataset of sentences annotated with \dialect{} labels to fine-tune one dialect-identification model for each language \cluster.


\item \textbf{Combined fine-tuning} ($\score{\mathcal{L}}{v}$): Unlike in the previous two methods, where each training set contains data from a single language \cluster, we fine-tune our baseline question-answering (EQA and MRC) models using the SD-QA~\cite{faisal-etal-2021-sd-qa} and Belebele~\cite{bandarkar2023belebele} datasets respectively, both of which contain training data in multiple standard varieties only and test data in other \dialects. The SD-QA training data ($\mathcal{L}$ in the notation) contains questions in 9 standard \dialects{} ($\mathcal{L}=\{\text{eng, ara, ben, fin, ind, swa, kor, rus, tel}\}$), while Belebele assembles data from 6 distinct multiple-choice QA datasets in standard English ($\mathcal{L}=\{\text{eng}\}$). 

\begin{table}[t]
\tiny
    \centering
    \begin{tabular}{@{}l|p{.9cm}@{ }p{.9cm}@{ }p{1cm}@{ }p{.5cm}@{ }p{.8cm}@{ }p{.9cm}@{}}
    \toprule
        Task & In-\dialect{} FT& In-\cluster{} FT & Combined FT & Zero-shot & No ref-erence & In-context learning\\
        \midrule
         DEP& $\checkmark$ &&&$\checkmark$&& \\
         POS & $\checkmark$ &&&$\checkmark$&& \\
         NER &  &$\checkmark$&&$\checkmark$&&\\
         
 EQA&  && $\checkmark$& $\checkmark$&&$\checkmark$\\
 MRC&  && $\checkmark$& &&\\
  NLI&  && & $\checkmark$&&\\
 TC&  &$\checkmark$& & $\checkmark$&&\\
  SA&  &$\checkmark$& & &&$\checkmark$\\
 DId&  &$\checkmark$& & &&\\
  MT&  && & $\checkmark$ &$\checkmark$ &\\
 \bottomrule
    \end{tabular}
    \caption{Task-specific training and evaluation procedure.}
    \label{tab:training}
\end{table}
\item \textbf{Zero-shot evaluation} ($\score{\text{eng}}{v}$): For certain \dialects{}, obtaining training data even for in-\cluster{} fine-tuning can be a challenge. Fortunately, English training data is always available for the datasets we study, so we use English to fill the gaps when we lack in-\dialect{}, in-\cluster{}, or combined training data. At the same time, we aim to assess the feasibility of using this zero-shot cross-lingual transfer in reducing any existing performance gap across \dialects. So we perform zero-shot cross-lingual transfer from English to each \dialect{} for 6 tasks in total. We only leave out those tasks such as dialect identification that explicitly require in-\cluster{} training data.

\item \textbf{In-context learning} ($\score{\text{icl}}{v}$): When evaluating large language models, we do not fine-tune them but instead rely on prompting and in-context learning. For this, we provide instructions and 5 examples in English as exemplars, followed by a prompt for predicting the test examples. Employing Mistral 7B~\cite{jiang2023mistral}, we assess the present effectiveness of a close-to-state-of-the-art LLM on language \dialects{}. The task-specific example prompts are reported in \cref{sec:icl_prompts}.
\end{compactenum}
\Cref{tab:training} summarizes the task-specific training procedures that we employ based on data availability.
Note that, for MT, we perform zero-shot evaluation specifically in the translation direction, (\textit{standard variety to English}) tested on (\textit{dialectal variety to English}). 
But evaluation is a challenge because human-created reference translations into or out of non-standard \dialects{} are usually very limited. Therefore, we adopt an evaluation protocol from previous work 
\citep{alam2023codet} that uses pseudo-references. Given $x$, a sentence in a \dialect{} and $\std{x}$, the translation of $x$ into the standard variety, let $y$ be the output of the MT system on input $x$ and $\std{y}$ be the output on input $\std{x}$. Then we measure the quality of $y$ (using, e.g., BLEU) compared against $\std{y}$ as a pseudo-reference.

\subsection{Quantifying the Dialectal Gap}\label{subsec:gap_metrics}

To quantify the performance disparity across various resource-specific settings, language \clusters{} and \dialects{}, we introduce a \emph{dialect performance gap} metric $\gapiii{t}{u}{v}$: the relative decrease in performance of a system fine-tuned on \dialect{}~$t$, tested on \dialect{} $v$ compared to a baseline \dialect{} $u$:
\begin{align*}
\gapiii{t}{u}{v} = \frac{\score{t}{u} - \score{t}{v}}{\score{t}{u}} \\
\intertext{with a special case for in-\dialect{} fine-tuning:}
\gapiii{\text{in-\dialect}}{u}{v} = \frac{\score{u}{u} - \score{v}{v}}{\score{u}{u}}.
\end{align*}
For the baseline score, we use either the score on the standard \dialect{} for each \cluster{} ($u=\std{v}$), or, in the zero-shot setting, the score on the language used for fine-tuning, namely English ($u=t=\text{eng}$). Rather than computing an absolute gap, we opt for a relative gap (i.e., dividing by the baseline score).
We also indicate whether the training setting is zero-shot ($t=\text{eng}$) or fine-tuned on in-\dialect, in-\cluster{}, or assembled data.
Putting all these together, we compute the following three variations of dialectal inequality.

\begin{compactenum}
\item $\gapiii{\text{eng}}{\text{eng}}{v}$: We calculate this metric to get a comprehensive measurement of global disparity across all \dialects{} in a resource-limited environment (zero-shot transfer from English).

\item $\gapiii{\text{eng}}{\std{v}}{v}$: Using this variation, we keep the setting fixed as zero-shot and calculate the gap between the \repr{} \dialect{} and any other \dialect{}. 

\item $\gapiii{t}{\std{v}}{v}$: The two aforementioned metrics shed light on the extent of the \dialect{} performance gap in a resource-limited setting. To gain a more comprehensive perspective, we additionally compute another metric, this time utilizing the availability of resources. The computation approach remains as straightforward as before. We just use fine-tuning on a \dialect{} $t$ instead of zero-shot transfer from Standard English: $t=\text{in-\dialect}$ for in-\dialect{} fine-tuning, $t=\std{v}$ for in-\cluster{} fine-tuning, or $t$ is some set of \dialects{} for combined fine-tuning.

\end{compactenum}
For all three $\gap$ metrics, we compute them at the \dialect{} level and then average them at the \cluster{} level:
\begin{align*}
  \gapiii{t}{u}{C} &= \frac1{|C|} \sum_{v \in C} \gapiii{t}{u}{v} \\
  \gapiii{\text{in-\dialect}}{u}{C} &= \frac1{|C|} \sum_{v \in C} \gapiii{\text{in-\dialect}}{u}{v}.
\end{align*}

\section{Results}
\begin{table*}
\centering
\tiny
\begin{tabular}{@{}llr@{ }rrrrll}
\toprule
&& num. & num. & avg. &&&&\\
Category           & Task   & \clus & \dial & score &        \multicolumn{2}{c}{Max-score \cluster/\dialect{}}  &        \multicolumn{2}{c}{Min-score \cluster/\dialect{}}    \\
\midrule
\multirow{3}{*}{Structured prediction} & DEP parsing &  16 &  40 &         64.3 &        sw. shifted romance/brazilian portuguese &         94.4 &    tupi-guarani sg./mbyá guaraní (a:brazil) &          9.0 \\
           & POS tagging &  17 &  51 &         72.1 &                   norwegian/norwegian bokmål (m:written) &         98.7 &    tupi-guarani sg./mbyá guaraní (a:brazil) &          1.9 \\
           & NER &  27 &  85 &         70.1 &                                 eastern romance/romanian &         94.2 &                       anglic/jamaican creole english &          0.0 \\
\cmidrule{1-9}
\multirow{4}{*}{Sequence classification} & NLI &  15 &  38 &         64.2 &                                           anglic/english &         83.4 &                   sotho-tswana (s.30)/southern sotho &         34.6 \\
           & TC &  15 &  38 &         77.7 &  sinitic/cmm. sinitic (o:traditional) &         89.8 &                              kurdish/central kurdish &         19.4 \\
           & DId &   6 &  49 &         67.0 &        sinitic/mandarin chinese (a:taiwan, o:simp.) &         98.6 &  sw. shifted romance/portuguese (m:written) &         17.4 \\
           & SA &   1 &   9 &         80.3 &                                   arabic/tunisian arabic &         94.6 &                        arabic/south levantine arabic &         58.9 \\
\cmidrule{1-9}
\multirow{2}{*}{Question Answering} & MRC &   4 &  11 &         40.9 &                                           anglic/english &         53.4 &                   sotho-tswana (s.30)/southern sotho &         29.0 \\
           & EQA &   5 &  24 &         74.2 &                           arabic/arabic (a:saudi-arabia) &         77.9 &                         swahili/swahili (a:tanzania) &         63.5 \\
\cmidrule{1-9}
\multirow{2}{*}{Generation} & MT-dialect &  12 &  73 &         25.2 &                               arabic/gulf arabic (a:riy) &         43.1 &                                  common turkic/sakha &          2.5 \\
           & MT-region &   2 &  41 &         33.0 &                     high german/central alemannic (a:ur) &         44.1 &                 italian romance/italian (a:sardegna) &         13.0 \\
\bottomrule
\end{tabular}
\caption{Task specific result summary using Maximum Obtainable Score. The \dialects{} with the minimum scores exhibit a noticeable lag in performance across various tasks when compared to the average task performance.}
\label{tab:comp_stat_main}
\end{table*}

We, first of all, discuss results by highlighting the highest possible score per \dialect{}, aka the \textit{maximum obtainable evaluation scores} regardless of evaluation method or training data. 
Next, we extend our discussion further by reporting the existing dialectal disparity across \clusters{} and \dialects{}.


\subsection{Maximum Obtainable Scores}
Here we provide key findings from our evaluation on each task.
A task-specific summary is reported in \cref{tab:comp_stat_main}. Detailed results comprising all tasks, models, language \clusters{} and \dialects{} are reported in \crefrange{tab:dep_full}{zero-shot-mt} in \cref{app:detailed_result_tables}. 


\begin{figure*}[!t]
       \centering
       \includegraphics[width=.9\textwidth]{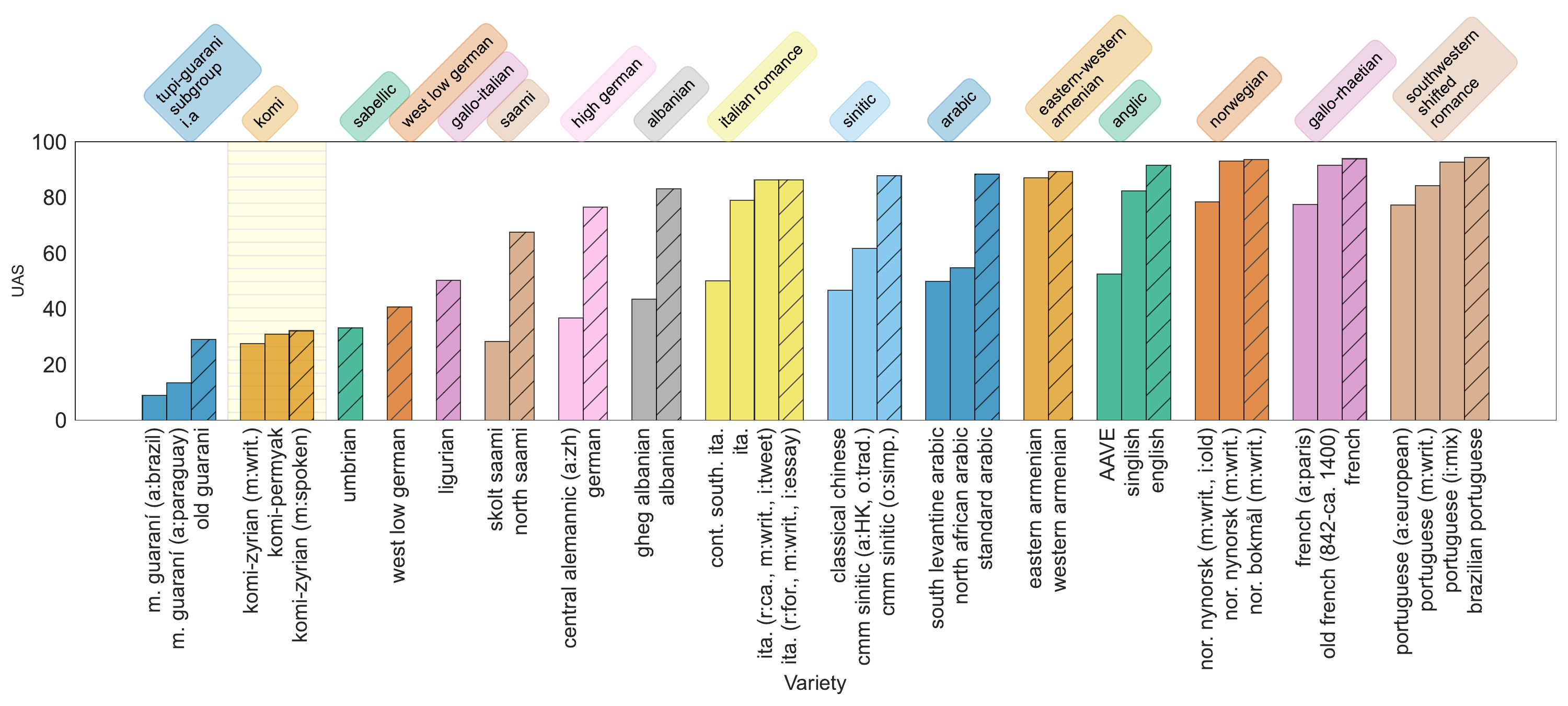}
       \label{fig:dep}
       \vspace{-1em}
    \caption{Maximum scores (\textit{max. UAS}) in Dep. Parsing task. \textit{Yellow-shaded region:} Komi is the only \cluster{} having no \dialects{} seen during mBERT pertaining. \textit{Colored bars with diagonal stripes:} the \cluster{} \repr{} \dialect{}. Low-resourced \cluster{} \dialects{} score lower compared to high-resource Germanic \clusters.}

    \label{fig:dep_plot}
\end{figure*}

\paragraph{Structured prediction} 
We present visualizations for the task-specific maximum scores. We show the one for Dependency Parsing in \cref{fig:dep_plot}, where we observe that low-resource varieties from language clusters such as Tupi-Guarani (indigenous South American \cluster{}), Saami and Komi (low-resource Uralic language \clusters{}) have the lowest performance compared to Standard English and other closely related Germanic and Romance \clusters{}. These low-resource varieties are also not included in the pretraining stage of our base language models (eg. mBERT).  Furthermore, this trend is evident across all three structured prediction tasks. On the other hand, high-resource Indo-European languages such as Portuguese, French, and Norwegian usually perform better. 
\paragraph{Sequence classification}

For DId and SA, we generally collate different datasets for each language \cluster{} and therefore, report the comparative classification results together. As a result, the locality level (e.g. city/region/country) also varies across \clusters{}. 
For example, we report city-level DId results for Arabic and High German but country-level results for Portuguese, Spanish and English. In the case of SA, we have region/country-level results for Arabic \dialects{}. For TC and NLI, we have the same set of \clusters{} and \dialects{}. However, we only report the zero-shot transfer performance from Standard English for NLI using the newly created translate-test NLI dataset. 

For TC and NLI, we observe the largest in-cluster disparity in the Kurdish cluster, with Northern Kurdish outperforming all others. The Sotho varieties consistently perform significantly lower compared to other \clusters{}. 
For all three sequence classification tasks, we generally find the Chinese \cluster{} performing on par with high-resource Latin counterparts. 


\paragraph{Question answering}
We generally do not see large gaps in performance within varieties in each language \cluster{}. In EQA zero-shot experiments, English and its \dialects{} have the highest performance overall and Korean \dialects{} score the lowest. Combined fine-tuning boosts performance on all language \clusters{} except in English. It's important to note that these EQA scores primarily indicate the model's robustness to accent-level differences and transcription noise, rather than broad dialectal robustness. However, further investigation is needed to determine whether this robustness specifically applies to both accent-level differences and transcription noise, or to any character-level variation up to a certain threshold.

For the MRC task, the performance peaked at 53.4 for Standard English, while the lowest score was 29 for Southern Sotho. More detailed results are presented in \cref{tab:mrc,tab:sdqa_test}. 


\paragraph{Machine translation}

The performance gap here varies widely across and within language varieties. Performance is similar within the Swiss-German \cluster{}, with higher performance (see \autoref{fig:swiss_german_mt})
across regions in Northern Switzerland, which is geographically closer to Germany. 
The performance gap for Norwegian dialects (Figure \ref{fig:mt-dialect}) is surprising as we perform zero-shot transfer from Norwegian Nynorsk (a Western dialect) but obtain better performance on the Eastern dialect.
Within Arabic, Riyadh is the highest performer while Sfax performs the worst. For the Bengali \cluster{}, Jessore has the highest performance --this is not surprising since it is one of the dialects from which standard Bengali originated \cite{alam2023codet}. The Ethiopian variety of Tigrinya exhibits a higher performance than the Eritrean one, even though Tigrinya is more commonly spoken in Eritrea \footnote{\url{https://en.wikipedia.org/wiki/Tigrinya_language}}. 
Amongst the \clusters{} within the Basque cluster, Barkoxe and Maule have the lowest score while Azkaine scores the highest.

\subsection{Dialectal Gap Across Language \Clusters{}}
\begin{figure*}[!t]
\begin{subfigure}[b]{0.33\textwidth}
    \centering
    \includegraphics[width=.8\linewidth]{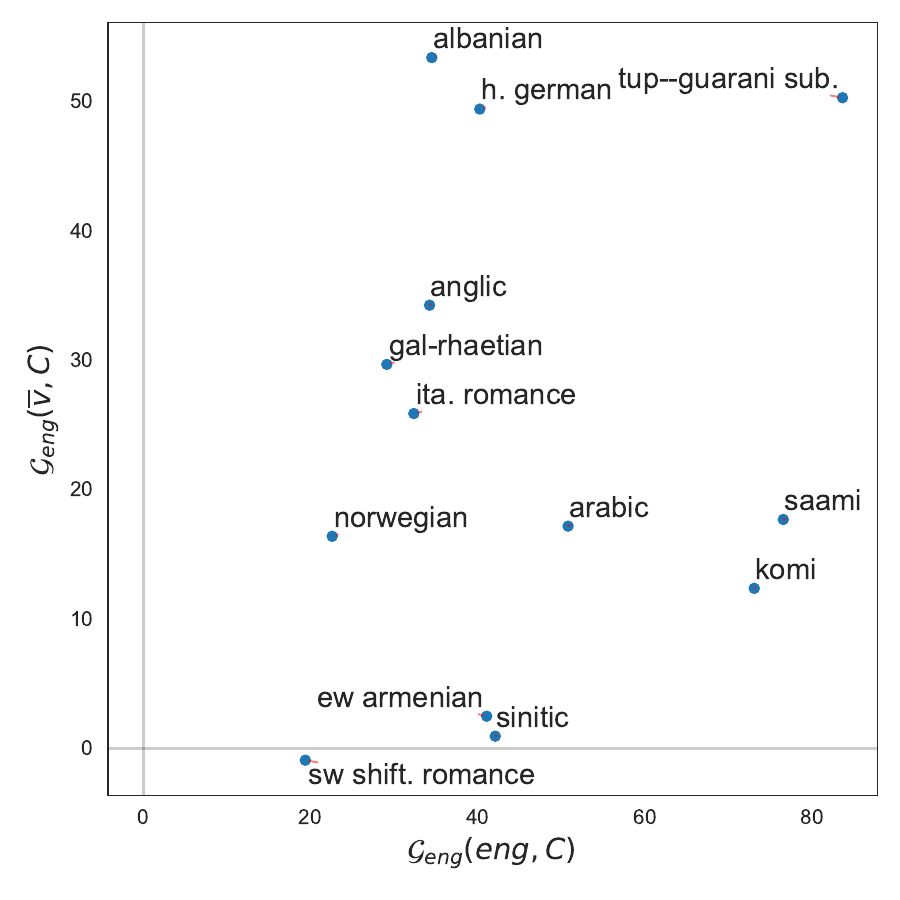}
    \caption{Dependency parsing}
    \label{fig:dep-gap}
\end{subfigure}
\begin{subfigure}[b]{0.33\textwidth}
    \centering
    \includegraphics[width=.8\linewidth]{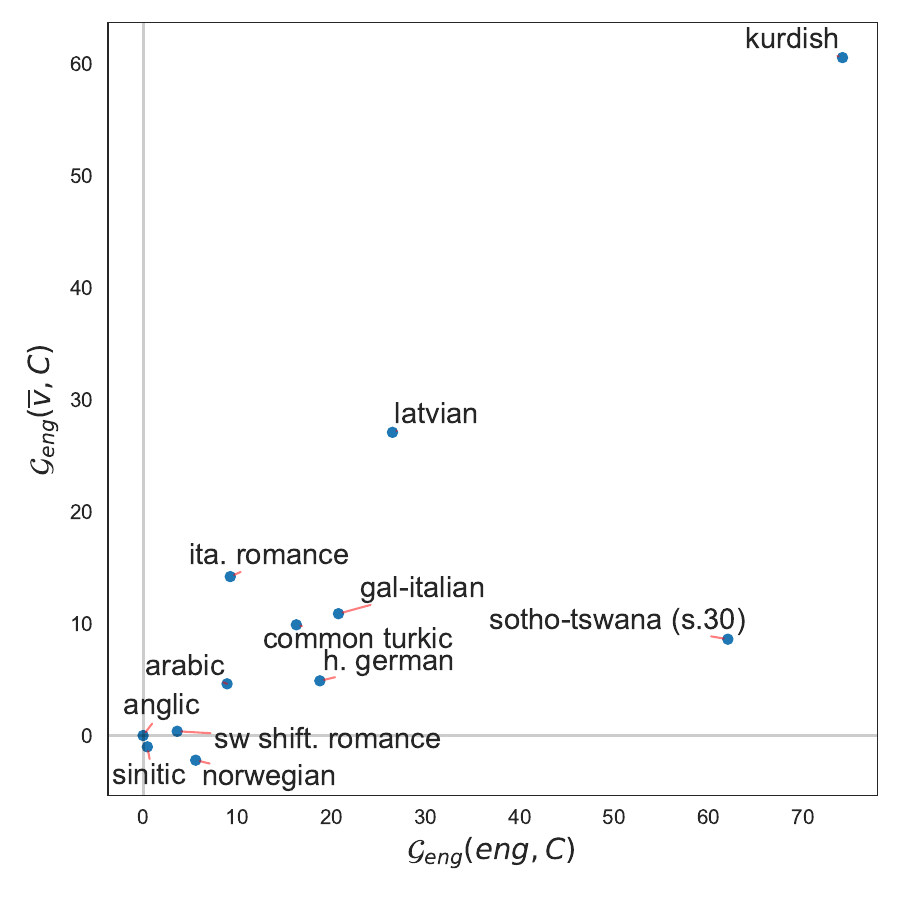}
    \caption{Topic classification}
    \label{fig:topic-gap}
\end{subfigure}
\begin{subfigure}[b]{0.33\textwidth}
    \centering
    \includegraphics[width=.8\linewidth]{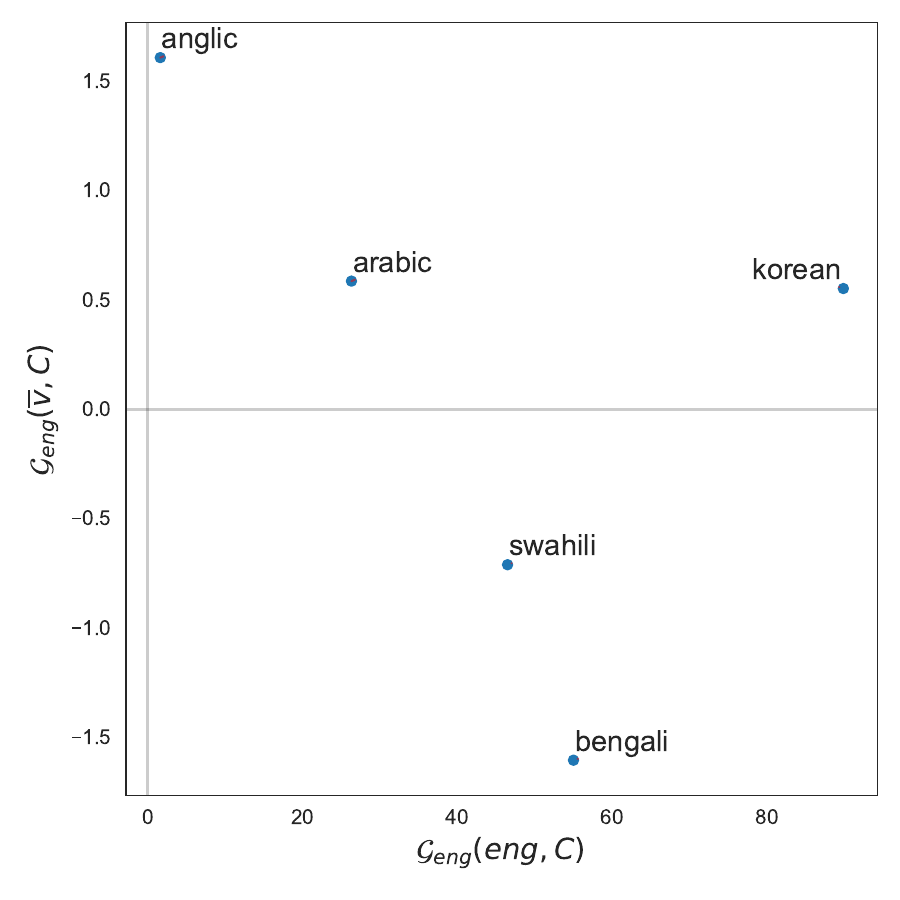}
    \caption{Extractive question answering}
    \label{fig:sdqa-test-gap}
\end{subfigure}
\caption{Dialectal gap visualization for language \clusters{} utilizing zero-shot cross-lingual transfer from Standard English. In the x-axis, values far from zero have a larger performance gap from English whereas, in the y-axis, values far from zero have a larger within cluster gap. Ideally, we want both of them to be close to zero. }
\vspace{-1em}
\label{fig:gap-plots}
\end{figure*}
 In \cref{fig:gap-plots}, we plot the zero-shot dialectal gap for three tasks. In the $x$-axis, we report the aggregated cluster-level gap $\gapiii{\text{eng}}{\text{eng}}{v}$, compared against the fine-tuning \dialect{} (standard English) while in $y$-axis we report $\gapiii{\text{eng}}{\std{v}}{v}$, the gap compared against the \repr{} \dialect{} of a \cluster{}. In an ideal scenario, we would want both of these gap values to be close to zero. However, this is certainly not the case. The general observed trend is that the low-resource \clusters{} have higher gaps of both $\gapiii{\text{eng}}{\text{eng}}{C}$ and \gapiii{\text{eng}}{\std{v}}{C}, whereas high-resource Germanic and Sinitic language \clusters{} consistently exhibit low dialectal gaps. That said, certain specific high-resource \dialects{}, such as Standard German and its dialectal counterparts like Swiss German, showcase significant within-\cluster{} dialectal gaps (\cref{fig:dep-gap}). 

\begin{table}[t]
\centering
\tiny
\begin{tabular}{@{}p{.8cm}l@{}r|r|r}
\toprule
Task & Gap metric                         &  Avg. val &             \cluster{} (max)&               \cluster{} (min)\\
\midrule
\multirow{3}{*}{DEP parsing} & $\gapiii{\text{eng}}{\text{eng}}{C}$ &      34.0 &   arabic, 50.8 &  sw. shift. romance, 19.4 \\
      & $\gapiii{\text{eng}}{\std{v}}{C}$ &      15.7 &   anglic, 34.2 &  sw. shift. romance, $-0.9$ \\
      & $\gapiii{\text{in-variety}}{\std{v}}{C}$ &      26.4 &   arabic, 93.8 &                italian romance, 0.6 \\
\cmidrule{1-5}
\multirow{3}{*}{POS tagging} & $\gapiii{\text{eng}}{\text{eng}}{C}$ &      27.4 &   arabic, 58.5 &                     norwegian, 14.7 \\
      & $\gapiii{\text{eng}}{\std{v}}{C}$ &       6.7 &   anglic, 20.9 &      ew. armenian, $-2.1$ \\
      & $\gapiii{\text{in-variety}}{\std{v}}{C}$ &       6.2 &   arabic, 29.1 &                          neva, $-0.5$ \\
\cmidrule{1-5}
\multirow{3}{*}{NER} & $\gapiii{\text{eng}}{\text{eng}}{C}$ &      31.7 &  kurdish, 77.1 &                  modern dutch, 12.3 \\
      & $\gapiii{\text{eng}}{\std{v}}{C}$ &      22.3 &  kurdish, 78.2 &                   hindustani, $-23.6$ \\
      & $\gapiii{\std{v}}{\std{v}}{C}$ &     $-28.6$ &  kurdish, 91.5 &                    sorbian, -1162.7 \\
\cmidrule{1-5}
\multirow{3}{*}{TC} & $\gapiii{\text{eng}}{\text{eng}}{C}$ &      22.4 &  kurdish, 74.2 &                        sinitic, 0.4 \\
      & $\gapiii{\text{eng}}{\std{v}}{C}$ &      12.5 &  kurdish, 60.6 &                     norwegian, $-2.2$ \\
      & $\gapiii{\std{v}}{\std{v}}{C}$ &      $-1.9$ &  latvian, 21.0 &                      kurdish, $-61.3$ \\
\bottomrule
\end{tabular}
\caption{Comparative cluster-level dialectal gap across tasks. In general, the average disparity is larger for zero-shot transfer $\gapiii{\text{eng}}{\text{eng}}{C}$. However, when we 
move from zeroshot to finetune (i.e. $\mathcal{G}_{\text{eng}}\rightarrow\mathcal{G}_{\bar{v}/\text{in-variety}}$) and compute the distance from a \cluster{} representative $\std{v}$, we observe increased dialectal disparity $|\gapiii{}{\std{v}}{C}|$.}
\label{tab:compare_gap}
\end{table}
We primarily report dialectal gaps using zero-shot transfer because the finetuning data available across task and \cluster{} is very disproportionate. Often the in-\cluster/\dialect{} data is not good enough in terms of data quality and quantity. For example, we have 37 \dialects{} in 13 \clusters{} for dependency parsing but out of these, only 20 \dialects{} have data available for in-\dialect{} fine-tuning. This lacking becomes more apparent when we compare the statistics of two types of within-cluster dialectal gaps: zero-shot \gapiii{\text{eng}}{\std{v}}{C} against fine-tuning \gapiii{\bar{v}/\text{in-variety}}{\std{v}}{C} in \cref{tab:compare_gap}. In general, the within-\cluster{} dialectal disparity is smaller for zero-shot transfer (i.e. $\gapiii{\text{eng}}{\std{v}}{C} \leq |\gapiii{\bar{v}/\text{in-variety}}{\std{v}}{C}|$). Here, the in-\cluster/\dialect{} fine-tuning results in a higher performance deviation primarily due to the inconsistent \dialect-specific finetuning data quality.



\section{Discussion}
\label{sec:discussion}
\paragraph{High resource vs.~low resource \dialects}
 The highest-performing \dialects{} are mostly standard high-resource languages and a few high-resource dialects (Norwegian dialects) whereas, the majority of the lowest-performing language variants are low-resourced \dialects{}. This clear distinction of language varieties points towards the large existence of in-\cluster{} dialectal gaps. Furthermore, this finding correlates with language script differences. We observe that  77.2\% of top-10 \dialects{} in terms of maximum obtainable score are written with Latin script. Another finding is the performance instability of 
low-resource \dialects{} across tasks. For instance, Old Guarani performs better in DEP parsing whereas, Mbyá Guarani (Paraguay) surpasses it in POS tagging even though the dataset remains the same (i.e. UD). For more detailed comparisons, in Appendix \cref{tab:high-lang}, we report the top-10 highest and lowest-scoring \dialects{} across different tasks.
\begin{table}[t]
\tiny
    \centering
   \begin{tabular}{@{}l|rr|rr@{}}
    \toprule
    {} &  \multicolumn{2}{c|}{zero-shot}&   \multicolumn{2}{c}{few-shot / FT} \\
    Task &    mBERT &    XLM-R &    mBERT &    XLM-R \\
    \midrule
    DEP. Parsing       &       61.6 &       61.3 &       76.2 &       64.3 \\
POS Tagging       &       69.5 &       69.7 &       89.8 &       89.1 \\
NER       &       59.7 &       57.8 &       65.8 &       61.4 \\
NLI       &       56.9 &       62.5 &        --- &        --- \\
TC     &       72.3 &       71.4 &       73.1 &       68.9 \\
MRC      &        --- &        --- &       39.4 &       40.3 \\
EQA &       53.9 &       51.9 &       69.2 &       67.2 \\
SA        &        --- &        --- &       78.8 &       80.1 \\
DId        &        --- &        --- &       65.8 &       59.3 \\
    \midrule
    win       &  4/6 &  2/6 & 6/8  &  2/8 \\
    \bottomrule
    \end{tabular}
    \caption{Base model comparison. We found mBERT was easier to fine-tune using the default hyperparameter setting thus, resulting in a higher winning rate.}
    \label{tab:model}
\end{table}

\paragraph{Model hyperparameter tuning} 
In \cref{tab:model}, we compare the baseline multilingual models. mBERT was comparatively easier to train than XLM-R using the default learning rates reported in the earlier task experiments. Often for tasks such as MRC, we needed to tune hyperparameters (e.g. learning rate, max. sequence length) in case of XLM-R. However, once we identify a hyperparameter configuration that converges in a zero-shot setting, we use the same to train all the language-variant training for that specific task. As a result, for some low-resource languages, XLM-R does not converge compared to mBERT. For example, XLM-R dependency parsing UAS score for Norwegian-NynorskLIA is 56.08 (zeroshot) and 8.25 (in-\dialect{} FT) whereas, we get 78.39 for in-\dialect{} mBERT fine-tuning. We suspect this hyperparameter tuning issue is one of the contributing factors toward a lower winning rate of XLM-R in few-shot / fine-tuning settings. However, this could be improved further with an extensive parameter grid search and settings specifically tailored for each language \cluster{}.

\paragraph{Positive zero-shot transfer for Latin \dialects{}}
Low-resource \dialects{} written in Latin script receive greater benefit in zero-shot because of effective transfer from high-resource Standard English. 
With the presence of In-\cluster/\dialect{} finetuning data, we effectively diminish this script effect to some extent. For example in the Hindi cluster, Fiji performs better than its Latin non-standard counterparts with in-\cluster{} finetuning for NER (\cref{tab:ner_full}). In summary, if the standard \dialect{} of a \cluster{} is non-latin but high-resource, then the success rate of in-\cluster/\dialect{} fine-tuning tends to be higher. However, where all dialects are low-resource, Latin script \dialects{} utilizing zero-shot transfer, eventually surpass others in the performance hierarchy. As an example, we report the zero-shot NER instances where the low-resource \dialects{} perform better than the representative ones in Appendix \cref{tab:script} (most of these use Latin script). 


\paragraph{LLM evaluation via In-context learning} For SA and EQA tasks, we have in-context learning results using the Mistral7B LLM. Comparing the performance against zero-shot and fine-tuning using our encoder-based models, we find dialect performance of LLM is better than zero-shot transfer but falls behind the finetuned results. 
On top of that, data contamination~\cite{ahuja2023megaverse} during LLM evaluation is another existing issue while considering these few available dialectal resources. Creating translation-based comparable data might be a solution to perform a fair benchmarking of LLMs on low-resource \dialects{}.



\paragraph{Misinterpreting evaluation metrics}
We also report the \cluster-level population-weighted average (i.e. \demu) which rewards a system more when it provides increased \lingu{} (eg. raw F1 score) for \dialects{}, spoken by a larger population compared to \dialects{} spoken by a smaller population. Alone, this metric could be misleading if we consider the fact that the performance gap among all \dialects{} from a particular \cluster{} should be minimal. 
On the other hand, solely looking into the \lingu{} average does not give a clear picture either (e.g. often overshadows the larger performance deviation of certain \dialects{} having extreme scores). So for all \clusters{} and tasks, we report the \lingu{} average as well as the \demu{} average, the minimum score of a \cluster{}, and the standard deviation in \crefrange{tab:summery-DEP. Parsing}{tab:summery-MRC}.

\section{Conclusion}
We propose \benchmark{}, the first-ever inclusive Dialectal NLP benchmark reporting performance evaluation and disparity across standard and non-standard \dialects{}. 
This is one step towards the effort of bringing more and more language under the paradigm of NLP technology. We would like to further improve the benchmark, constructing high-quality comparable data and expanding the task coverage to speech-based NLP technologies.

\section*{Limitations}
The data quality and quantity, \dialect{} coverage vary significantly across tasks because of data scarcity issues. We avoid full-scale LLM-evaluation consciously because of the uncertain data-contamination issue and their well-known lower performance threshold compared to smaller masked-language-modeling-based fine-tuned models. In addition, we focus on text-based NLP tasks for this current iteration. Moreover, we do not claim the \repr{} \dialects{} of each language \clusters{} to be any kind of superior or standardized forms over the other \dialects{}. These \dialects{} are chosen to perform a well-informed comparison among the perceived well-resourced linguistic \dialect{} and its counterparts having lesser data availability. At the same time, the mutual intelligibility and phylogenetic similarity of the similar \cluster{} \dialects{} also vary across \cluster{} and this was not selected in a numerically quantifiable manner.

\paragraph{Evaluation Limitations}
\label{sec:improvement_scope}
To further improve the evaluation fairness of the current version of \benchmark{}, we need (i) Parallel corpus utilization to prepare task-specific data (ii) Translation-based task data generation to perform comparable analysis (iii) Quantifying the resource-supply and demand as well as population-coverage~\cite{song2023globalbench} to identify where a \dialect{} stands in the global landscape of \lingu{}. Here, we have accumulated data for diverse \dialects{} across tasks that vary significantly in terms of quality, example count, and domain. However, to perform a perfectly fair comparison of dialectal inequality, we should consider high-quality comparable data (e.g. parallel corpus, similar \dialects{} across tasks) which is not available at this point.

\paragraph{Continuity of \benchmark} Despite our best effort, this benchmark does not include every one of the already published task-specific dialectal datasets. So, our next steps on this project involve hosting the benchmark on the website that displays the current statistics of the datasets in \benchmark. We will also encourage researchers to add new and existing datasets for tasks, language \clusters{} that might be currently missing alongside the respective baselines.

\paragraph{Space Limitations} Our study encompasses a large set of evaluation result tables, their corresponding visualizations and findings analysis. Due to space limitations, we have to move the detailed reports (\cref{app:detailed_result_tables}) and the rest of the visualizations (\cref{app:plots}) in the Appendix. To better assist, we include an Appendix Table of Content (\cref{tab:app_content}) at the introductory section of Appendix (\cref{app:start}). 

\section*{Ethics Statement}
This work is a compilation of existent dialectal datasets across different tasks, including structured prediction and generative tasks. Our experiments do not particularly optimize for the best model performance of these tasks. Therefore we acknowledge that for some of the tasks, the baseline models might not be robust enough to handle dialectal text hence resulting in wrong predictions and generations. We believe that this underscores the need for building models robust to different language variations and future work should focus on this.


\section*{Acknowledgements}
This material is based upon work supported by the US National Science Foundation under Grants No. IIS-2125466, IIS-2125948, CAREER Grant No.~IIS2142739, as well as NSF Grants No.~IIS-2203097 and IIS-2125201. We gratefully acknowledge support from Alfred P.~Sloan Foundation Fellowship. This research is also supported in part by the Office of the Director of National Intelligence (ODNI), Intelligence Advanced Research Projects Activity (IARPA), via the HIATUS Program contract \#2022-22072200004. The views and conclusions contained herein are those of the authors and should not be interpreted as necessarily representing the official policies, either expressed or implied, of ODNI, IARPA, or the U.S. Government. The U.S. Government is authorized to reproduce and distribute reprints for governmental purposes notwithstanding any copyright annotation therein.

\bibliography{anthology,custom}
\bibliographystyle{acl_natbib}

\appendix
\onecolumn

\newpage
\section*{Appendix}
\label{app:start}
In this supplementary material, we provide the following: (i) Relevant literature review (ii) Overall results and dataset details 
that we could not fit into the main body of the paper.
\begin{table}[ht]
\small
    \begin{tabular}{l|l}
    \toprule
     Section & Desctiption \\
     \midrule
     \cref{app:rel_work} & Related Work \\
     \cmidrule{1-2}
     \multirow{2}{*}{\cref{sec:tasks_benchmark}}    & Tasks of \benchmark\\
     & \tabitem Translate-Test NLI Dataset Statistics (\cref{{tab:nli_data}}) \\

    \cmidrule{1-2}
      \multirow{3}{*}{\cref{app:lang_stdv}} &  \Dialects{} and \Clusters{} of \benchmark \\
     & \tabitem \benchmark{} \Dialect{} list (\cref{lang_list})\\
     & \tabitem Language \Clusters{} and Representative \Dialects{} (\cref{stdv})\\

     \cmidrule{1-2}
     \multirow{18}{*}{\cref{app:plots}} & Result Visualizations\\ \\
     & 
     
    \begin{tabular}{l}
        Regional maps with aggregated Machine Translation scores (\crefrange{fig:swiss_german_mt}{fig:italian_mt})\\
        \cmidrule{1-1}
         \tabitem Map of Switzerland with aggregated BLEU scores of Swiss-German variety per region (\cref{fig:swiss_german_mt}) \\
         \tabitem Map of Italy with aggregated BLEU scores of Italian variety per region (\cref{fig:italian_mt}) \\\\

         Task Specific Plot for Maximum Scores (\crefrange{fig:plot_app_1}{fig:plot_app_4})\\
        \cmidrule{1-1}
         \tabitem Parts-of-Speech Tagging (\cref{fig:pos}) \\
        \tabitem Named entity recognition (\cref{fig:ner})\\
        
        \tabitem Topic classification  (\cref{fig:tc})\\
        \tabitem Natural language inference (\cref{fig:nli})\\
        
        \tabitem Extractive question answering (\cref{fig:qa})\\
        \tabitem Multiple-choice machine reading comprehension (\cref{fig:rcmc})\\
        \tabitem Sentiment analysis (\cref{fig:sc})\\
        \tabitem Dialect identification (\cref{fig:di}) \\

        \tabitem Machine translation (MT-region) (\cref{fig:mt-region})\\
        \tabitem Machine translation (MT-dialect) (\cref{fig:mt-dialect})\\
        
        \\

        Dialectal Gap visualization  utilizing zero-shot cross-lingual transfer from Standard English.\\
        \cmidrule{1-1}
        \tabitem Parts-of-Speech Tagging (\cref{fig:pos-gap})\\
        \tabitem Named entity recognition (\cref{fig:ner-gap})\\
        \tabitem Dialect identification (\cref{fig:nli-gap})\\
    \end{tabular}\\

    \cmidrule{1-2}
     \multirow{11}{*}{\cref{app:detailed_result_tables}} & Task Specific Evaluation Result Tables\\ 
     & 
    \begin{tabular}{l}
        \cmidrule{1-1}
        \tabitem Dependency parsing (\cref{tab:dep_full}) \\
         \tabitem Parts-of-Speech tagging (\cref{tab:pos_full}) \\
        \tabitem Named entity recognition (\cref{tab:ner_full})\\
        \tabitem Natural language inference (\cref{tab:nli})\\
        \tabitem Extractive question answering (\crefrange{tab:sdqa_dev}{tab:sdqa_test})\\
        
        \tabitem Dialect identification (\cref{tab:id_all}) \\
        \tabitem Topic classification  (\cref{tab:topic})\\
        \tabitem Sentiment analysis (\cref{tab:sc})\\
        
        \tabitem Multiple-choice machine reading comprehension (\cref{tab:mrc})\\
        
        \tabitem Machine translation (\cref{zero-shot-mt})\\
    \end{tabular}\\

    \cmidrule{1-2}
    \cref{app:high_low} & Highest performing and lowest performing varieties (\cref{{tab:high-lang}}) \\

    \cmidrule{1-2}
    \cref{app:non_std_zeroshot} & Low-resource \Dialect{} performing better in zero-shot NER (\cref{tab:script}) \\
    
    \cmidrule{1-2}
     \cref{app:c_summary}    & \Cluster{} level Result Summaries with Demographic Utility and Standard Deviation report (\crefrange{tab:summery-DEP. Parsing}{tab:summery-EQA})\\

     \cmidrule{1-2}
     \cref{sec:icl_prompts} & In-Context Learning Details \\
     
    \bottomrule
    \end{tabular}
    \caption{Table of contents for supplementary material.}
    \label{tab:app_content}
\end{table}


\section{Related Work} 
\label{app:rel_work}
The majority of multilingual benchmarks \cite{pmlr-v119-hu20b, ruder2023xtremeup, ruder-etal-2021-xtreme, liang-etal-2020-xglue-fixed, wilie-etal-2020-indonlu, park2021klue} have heavily focused on dominant language varieties. However, the performance disparity between standard languages and their dialectal counterparts has been studied \cite{kantharuban2023quantifying, ziems-etal-2022-value, Rios_2020, ziems-etal-2022-multi-value} and shown to be significantly large across different tasks. Still, the large-scale generalization of this finding comprising numerous task categories is yet to be done. There have been previous efforts to bridge this gap though.  Some work has focused on curating dialectal datasets across several tasks within one language cluster, while others have focused on single tasks across many language clusters. For instance, ORCA \cite{elmadany-etal-2023-orca}, ARLUE \cite{abdul-mageed-etal-2021-arbert} and ALUE \cite{seelawi-etal-2021-alue} are dedicated natural language understanding benchmarks focusing on Arabic varieties alone. Multi-VALUE \cite{ziems-etal-2022-multi-value} was developed for benchmarking NLP tasks in English varieties. \cite{mirzakhalov-etal-2021-large} is a suite of resources for benchmarking MT in several Turkic languages. There has also been a large body of work on dialect identification across several languages \cite{jauhiainen-etal-2022-italian, hamalainen-etal-2021-finnish}. Recently CODET \cite{alam2023codet} was released as a contrastive dialectal MT benchmark  covering  882 different variations from nine different languages. To the best of our knowledge, we are the first to do a large scale aggregation of dialectal data across several language clusters and tasks.

\section{Tasks of \benchmark}\label{sec:tasks_benchmark}
\benchmark{} includes 10 NLP tasks falling into four broader categories: structured prediction, sentence classification, question answering, and text generation. In \cref{tab:task}, we present statistics for the datasets for each task and briefly discuss each task below. 

\paragraph{Dependency parsing} 
For the dependency parsing task, we include only those Universal Dependencies (UD)~\cite{11234/1-4611} languages that have dialectal data. 
Beyond the data available in UD 2.12, we incorporate two additional datasets for African-American English (AAVE) Twitter data~\cite{blodgett-etal-2018-twitter} and Singlish~\cite{wang-etal-2017-universal-fixed}. To make these two datasets compatible with the UD processing pipeline, we replace the original dependency labels with the labels corresponding to the official UD formalism. 

\paragraph{Part of speech (POS) tagging} We use the same UD languages for POS tagging that we used for dependency parsing. At the same time, we use the POS data instances from Singlish~\cite{wang-etal-2017-universal-fixed}. Moreover, we include six Finnish dialects, four Arabic dialects and Occitan through the UPOS label standardized pipeline proposed by~\citet{blaschke-etal-2023-manipulating}.

\paragraph{Named entity recognition (NER)} We use data from the 176-language version of the Wikiann dataset processed by~\citet{rahimi-etal-2019-massively}. All these languages provide both training and test data. In addition, we include dialectal data from the original Wikiann dataset (282 languages)~\cite{pan-etal-2017-cross} for evaluation. Moreover, we include three Norwegian dialects~\cite{johansen2019ner} with train, test and validation datasets that use a slightly different set of NER tags (GEO, ORG, OTH, PER) compared to the one we use in Wikiann (LOC, ORG, PER). We leave these levels as it is and do not convert to the Wikiann tags.

\paragraph{Dialect identification} 
We include dialect identification experiments on Arabic, Greek, Portuguese, English, Spanish, and Swiss German dialectal datasets. In these datasets, we find large variations in the level of granularity with which dialects are classified. For instance, the MADAR corpus differentiates Arabic \dialects{} at the city level~\cite{bouamor-etal-2018-madar}, whereas our English and Spanish datasets are labeled with country names~\cite{zampieri2023language}. 

\paragraph{Sentiment classification} Here we include several different Arabic \dialects{}. Like other dialectal datasets, these datasets do not follow one standard labeling process. However, all datasets contain two main sentiment types: positive and negative. A number of datasets contain additional labels such as ``objective'' or ``neutral.'' 
In our setting, we perform a further split of data to provide validation data for each dialect. However, we do not remove these extra labeled data for information preservation.

\paragraph{Topic classification} We use the SIB-200 dataset \cite{adelani2023sib200} for topic classification task 
SIB-200 was constructed from the FLORES-200 translation datasets. The authors annotated the English dataset of FLORES-200 with 6 topic labels and then further propagated these labels to the translated instances for all other languages. For our case of benchmarking dialectal segments, we use the dialectal and regional \dialects{} from SIB-200. 

\begin{table*}
\centering
\small
    \begin{tabular}{lllr}
\toprule
     &          &                    \Dialect{} &  \# Sentences \\
\cluster{} & Language code &                         &       \\
\midrule
anglic & eng\_Latn &                                          english &  5010 \\
\cmidrule{1-4}
\multirow{9}{*}{arabic} & acm\_Arab &                        north mesopotamian arabic &  5010 \\
                             & acq\_Arab &                             ta'izzi-adeni arabic &  5010 \\
                             & aeb\_Arab &                                  tunisian arabic &  5010 \\
                             & ajp\_Arab &                           south levantine arabic &  5010 \\
                             & apc\_Arab &                       levantine arabic (a:north) &  5010 \\
                             & arb\_Arab &                                  standard arabic &  5010 \\
                             & ars\_Arab &                                     najdi arabic &  5010 \\
                             & ary\_Arab &                                  moroccan arabic &  5010 \\
                             & arz\_Arab &                                  egyptian arabic &  5010 \\
\cmidrule{1-4}
\multirow{3}{*}{common turkic} & azb\_Arab &                                south azerbaijani &  5010 \\
                             & azj\_Latn &                                north azerbaijani &  5010 \\
                             & tur\_Latn &                         central oghuz (m:spoken) &  5010 \\
\cmidrule{1-4}
\multirow{3}{*}{gallo-italian} & lij\_Latn &                                         ligurian &  5010 \\
                             & lmo\_Latn &                                          lombard &  5010 \\
                             & vec\_Latn &                                         venetian &  5010 \\
\cmidrule{1-4}
gallo-rhaetian & fur\_Latn &                                         friulian &  5010 \\
\multirow{2}{*}{high german} & lim\_Latn &                                        limburgan &  5010 \\
                             & ltz\_Latn &                                     luxemburgish &  5010 \\
\cmidrule{1-4}
\multirow{2}{*}{italian romance} & ita\_Latn &                                          italian &  5010 \\
                             & scn\_Latn &                                         sicilian &  5010 \\
\cmidrule{1-4}
\multirow{2}{*}{kurdish} & ckb\_Arab &                                  central kurdish &  5010 \\
                             & kmr\_Latn &                                 northern kurdish &  5010 \\
\cmidrule{1-4}
\multirow{2}{*}{latvian} & ltg\_Latn &                                     east latvian &  5010 \\
                             & lvs\_Latn &                                          latvian &  5010 \\
\cmidrule{1-4}
modern dutch & nld\_Latn &                                            dutch &  5010 \\
\multirow{2}{*}{norwegian} & nno\_Latn &                    norwegian nynorsk (m:written) &  5010 \\
                             & nob\_Latn &                     norwegian bokmål (m:written) &  5010 \\
\cmidrule{1-4}
sardo-corsican & srd\_Latn &                                        sardinian &  5010 \\
\multirow{3}{*}{sinitic} & yue\_Hant &                                        cantonese &  5010 \\
                             & zho\_Hans &   classical-middle-modern sinitic (o:simplified) &  5010 \\
                             & zho\_Hant &  classical-middle-modern sinitic (o:traditional) &  5010 \\
\cmidrule{1-4}
\multirow{2}{*}{sotho-tswana (s.30)} & nso\_Latn &                                   northern sotho &  5010 \\
                             & sot\_Latn &                                   southern sotho &  5010 \\
\cmidrule{1-4}
\multirow{4}{*}{southwestern shifted romance} & glg\_Latn &                                         galician &  5010 \\
                             & oci\_Latn &                                          occitan &  5010 \\
                             & por\_Latn &                          portuguese (a:european) &  5010 \\
                             & spa\_Latn &                                          spanish &  5010 \\
\bottomrule
\end{tabular}
\caption{Data statistics for newly created translate-test natural language inference (NLI) dataset. We prepare this translate-test NLI dataset by translating XNLI~\cite{conneau-etal-2018-xnli} english evaluation dataset.}
\label{tab:nli_data}
\end{table*}

\paragraph{Natural language inference} For the natural language inference task, there is no existing dataset with \dialects. So we use the existing English test set of XNLI~\cite{conneau-etal-2018-xnli} and construct a multilingual dialect-focused translated evaluation dataset. We use a state-of-the-art machine translation model (NLLB-200 3B) to translate the English test set to 12 language \clusters{} encompassing 40 \dialects{} (Complete data statistics are reported in \cref{tab:nli_data}). After that, we perform zero-shot cross-lingual transfer from the English finetuned NLI model. We refer to this setting as \emph{translate-test}. 

\paragraph{Multiple-choice machine reading comprehension} This task aims to evaluate the capability of multiple-choice question answering given a context passage. The question could be answered by understanding the context passage while the right answer is given at one of the multiple choices. We use the Chinese, Sotho and Arabic clusters data from the recently released Belebele MRC dataset~\cite{bandarkar2023belebele}. This is an evaluation-only dataset.  

\paragraph{Extractive question answering} This task predicts the answer span given a question and context passage. We use the SD-QA dialectal question-answering dataset~\cite{faisal-etal-2021-sd-qa}. SD-QA is an evaluation dataset built on top of TyDiQA~\cite{clark-etal-2020-tydi}, another well-known typologically diverse question-answering dataset. SD-QA contains the spoken utterances and transcription of the original TyDiQA question from speakers of  English, Bengali, Arabic, Korean, and Swahili. We only use the textual part that contains the transcription of the dialectal spoken question matching the original TyDiQA question text. Note that since the transcriptions of the questions are obtained through automatic speech recognition, they may include both dialectal variations \textit{and} noise due to ASR transcription errors. 

\paragraph{Machine translation}  We evaluate \dialect{} translation using the CODET benchmark~\cite{alam2023codet} and the TIL MT corpus \cite{mirzakhalov2021large}. CODET contains a contrastive dataset of 882 different \dialects{} from nine different languages. We evaluate dialects here at city level for all languages except Italian and Swiss-German and 
aggregate dialects at the region level for them.
The TIL corpus contains parallel translations across 22 Turkic languages, but in our evaluations we only include 8 turkic languages (Turkic, Sakha, Kazakh, Karakalpak, Bashkir, Azerbaijani, Kyrgyz) 
that have parallel English translations.

\section{\Dialects{} and \Clusters{} of \benchmark}
\label{app:lang_stdv}
\subsection{\benchmark{} \Dialect{} list}
\label{app:lang}
\tiny


\normalsize
\subsection{Language \Clusters{} and Representative \Dialects{}}
\label{app:c_stdv}
\tiny
\begin{longtable}{lll}
\caption{Language \clusters{} and their standard \dialects.}
\label{stdv}\\
\toprule
Task & \Cluster{} Name &        \Cluster{} \Repr                           \\
\midrule
\endfirsthead
\caption[]{Language \clusters{} and their \clusters{} representatives.} \\
\toprule
Task & \Cluster{} Name &        \Cluster{} \Repr                           \\
\midrule
\endhead
\midrule
\multicolumn{3}{r}{{Continued on next page}} \\
\midrule
\endfoot

\bottomrule
\endlastfoot
\multirow{13}{*}{DEP. Parsing} & albanian &                                        albanian \\
    & arabic &                                 standard arabic \\
    & eastern-western armenian &                                western armenian \\
    & sinitic &  classical-middle-modern sinitic (o:simplified) \\
    & anglic &                                         english \\
    & gallo-rhaetian &                                          french \\
    & high german &                                          german \\
    & tupi-guarani subgroup i.a &                                     old guarani \\
\end{longtable}

\normalsize

\section{Result Visualizations}
\label{app:plots}
\subsection{Regional maps with aggregated Machine Translation scores}
\begin{figure}[ht]

\includegraphics[width=.5\columnwidth]{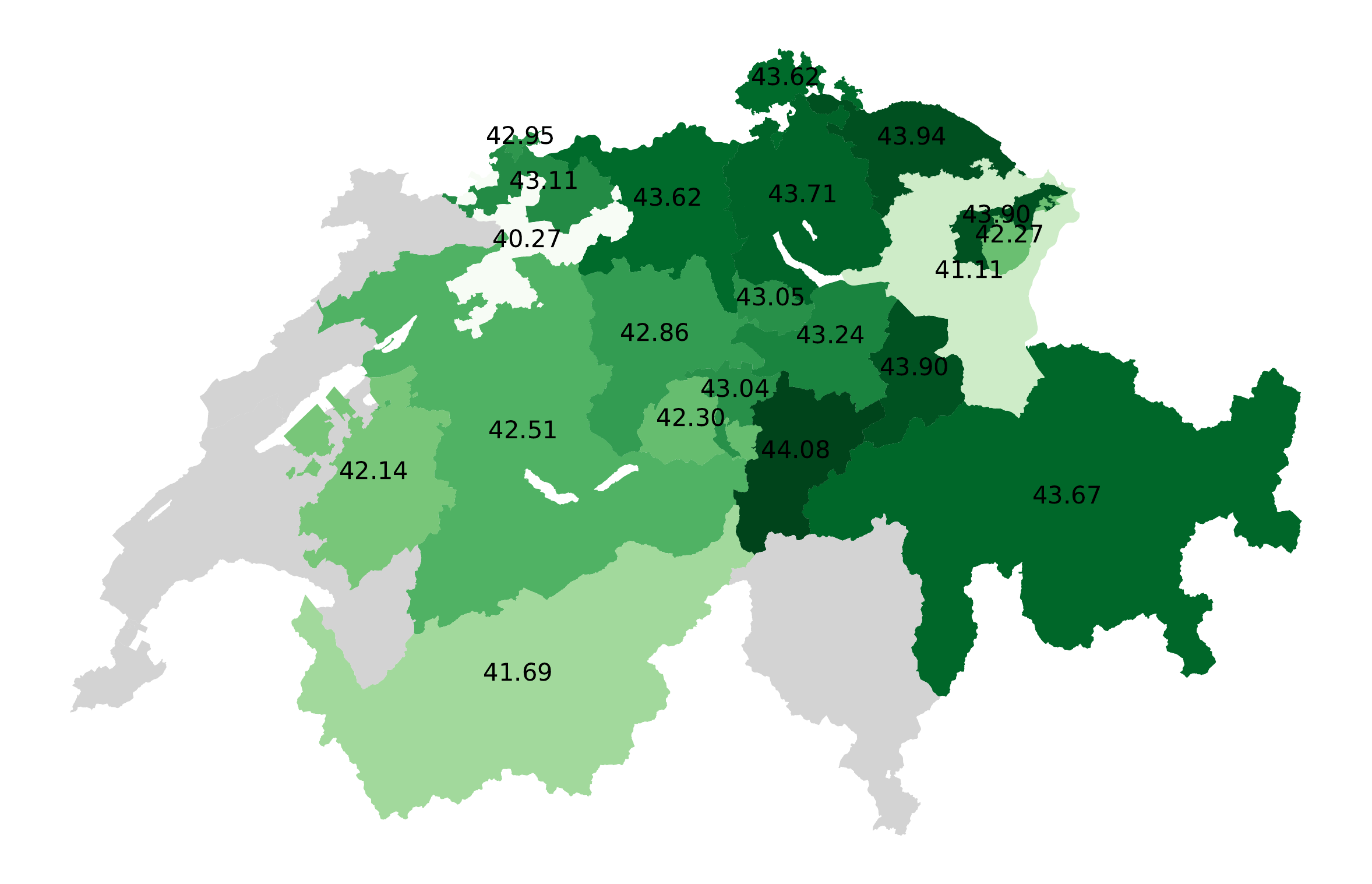}
 \caption{Map of Switzerland with aggregated BLEU scores of Swiss-German \dialect{} per region}
    \label{fig:swiss_german_mt}
\end{figure}

\begin{figure}[ht]
\includegraphics[width=0.5\textwidth]{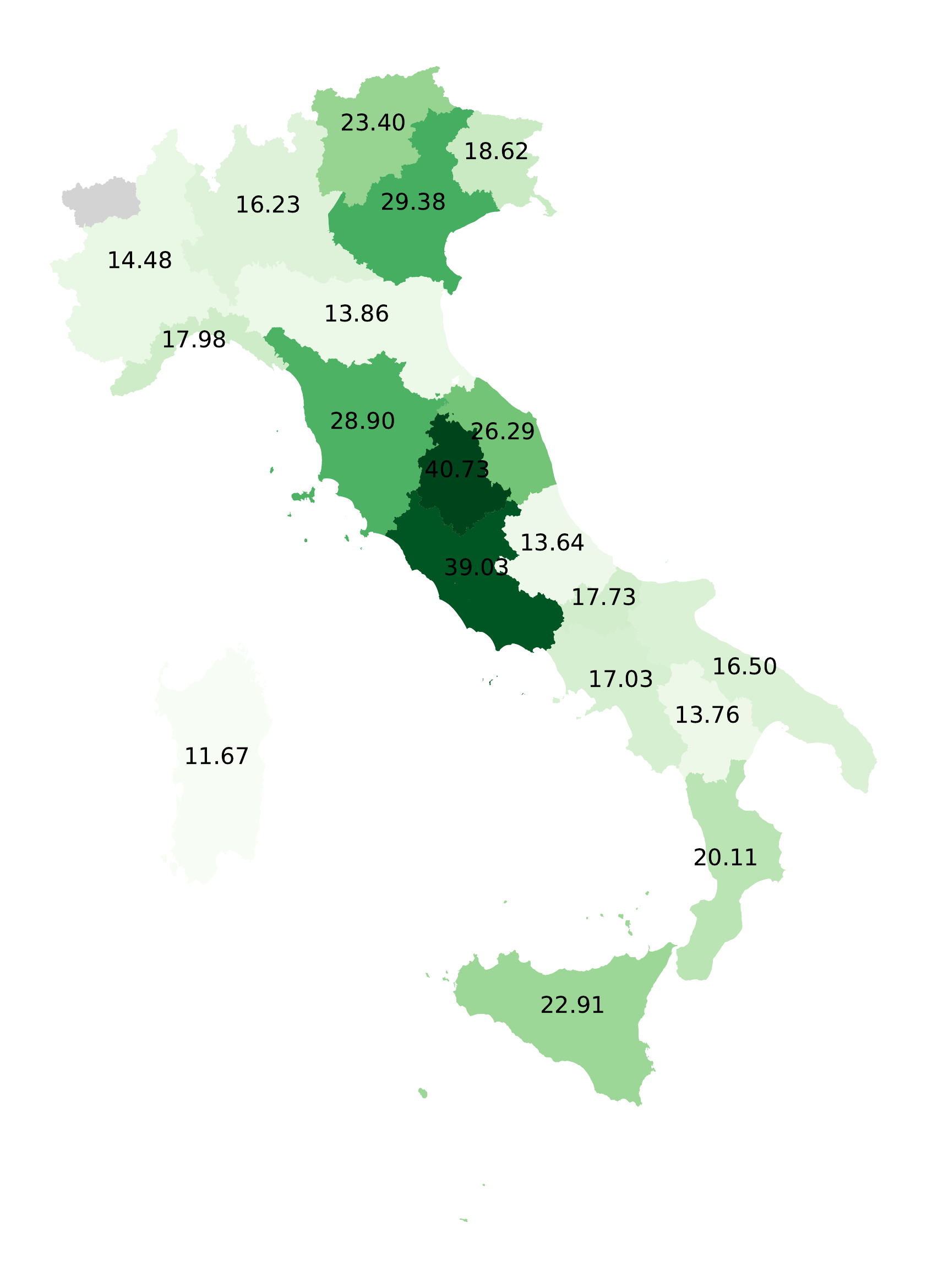}
 \caption{Map of Italy with aggregated BLEU scores of Italian \dialect{} per region}
    \label{fig:italian_mt}
\end{figure}
\subsection{Task Specific Plot for Maximum Scores}
\begin{figure*}[ht]
    \begin{subfigure}{\linewidth}
    \centering
    \includegraphics[width=.9\textwidth]{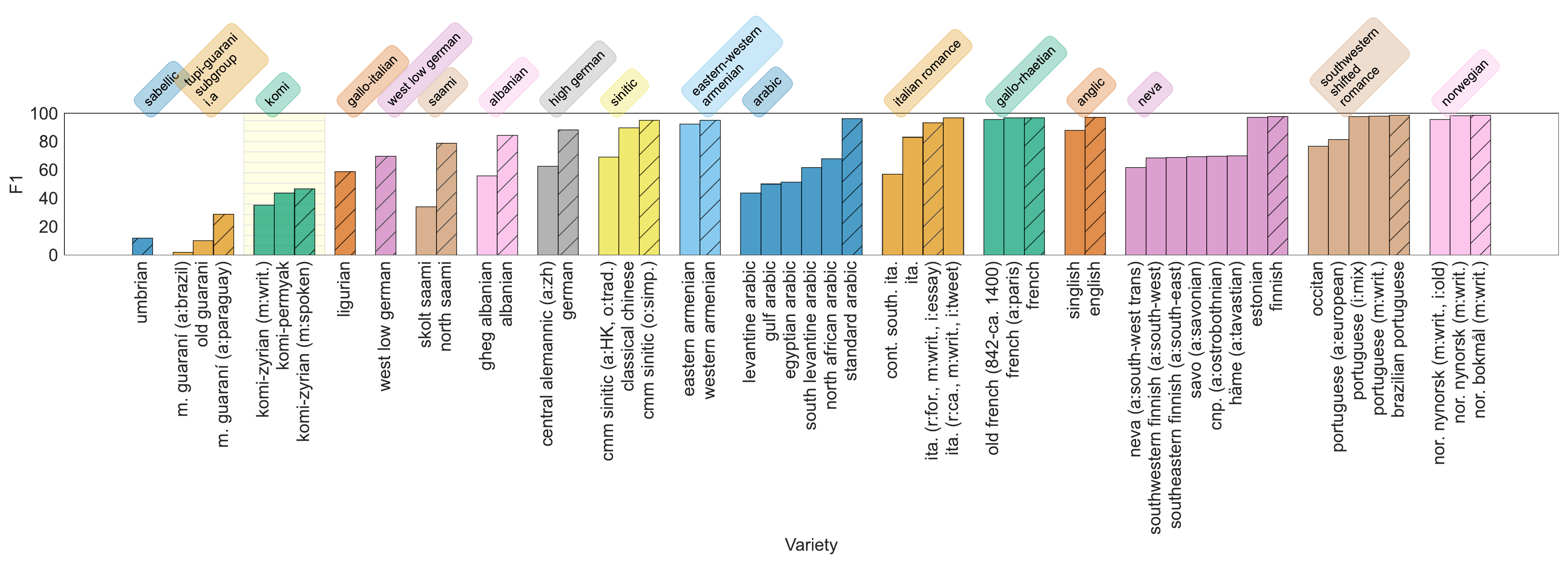}
    \caption{Parts of Speech Tagging} 
    \label{fig:pos}
    \end{subfigure}
    \begin{subfigure}{\linewidth}
    \centering
    \includegraphics[width=1\textwidth]{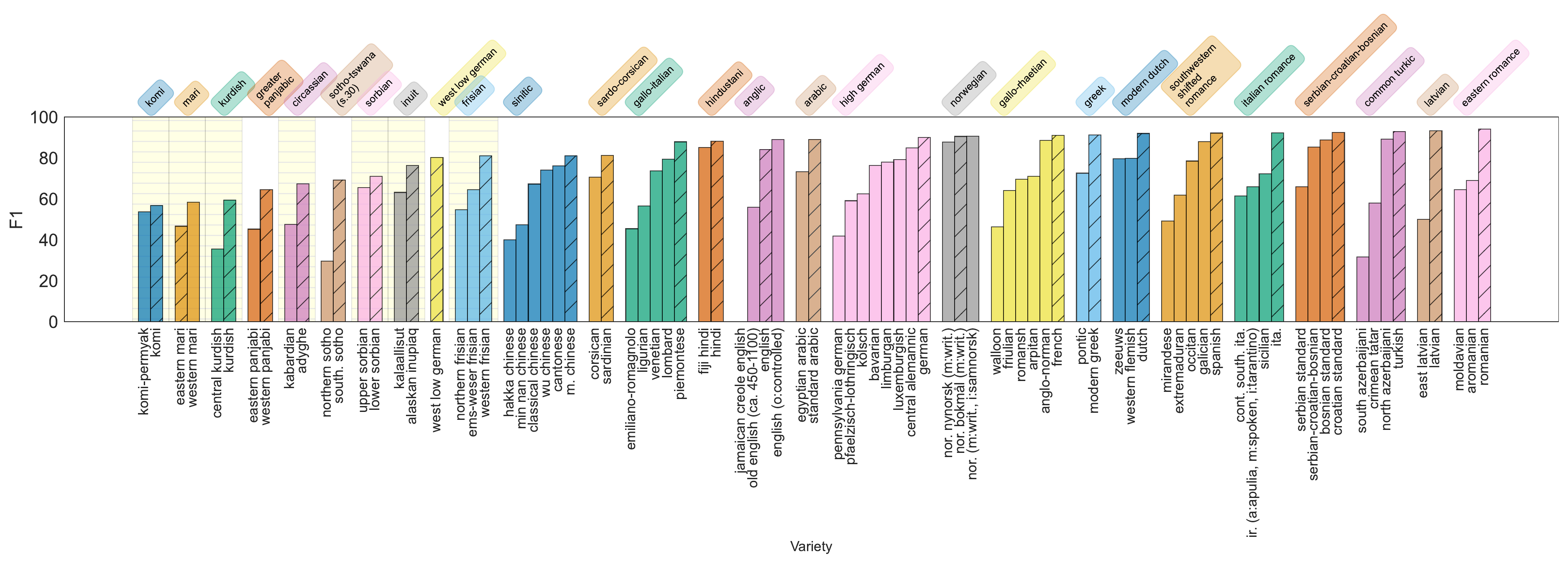}
    \caption{Named Entity Recognition}
    \label{fig:ner}
    \end{subfigure}

    \begin{subfigure}{\linewidth}
    \centering
    \includegraphics[width=1\textwidth]{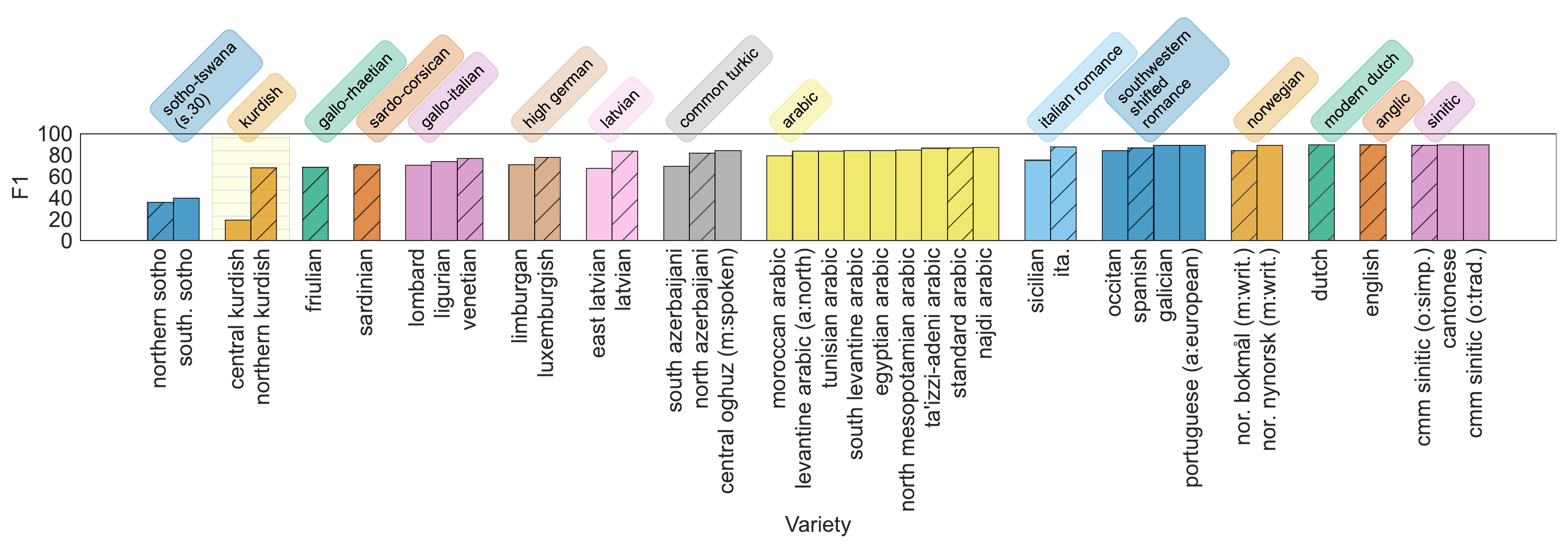}
    \caption{Topic Classification}
    \label{fig:tc}
    \end{subfigure}

    \caption{Task specific plot of maximum obtainable score for all \dialects{}. The yellow-shaded region represents language \clusters{} having no \dialects{} seen during mBERT pertaining. The bars with colored stripes represent the standard \dialect{} of a \cluster{}}
    \label{fig:plot_app_1}

\end{figure*}

\begin{figure*}[ht]
    \begin{subfigure}{\linewidth}
    \centering
    \includegraphics[width=.9\textwidth]{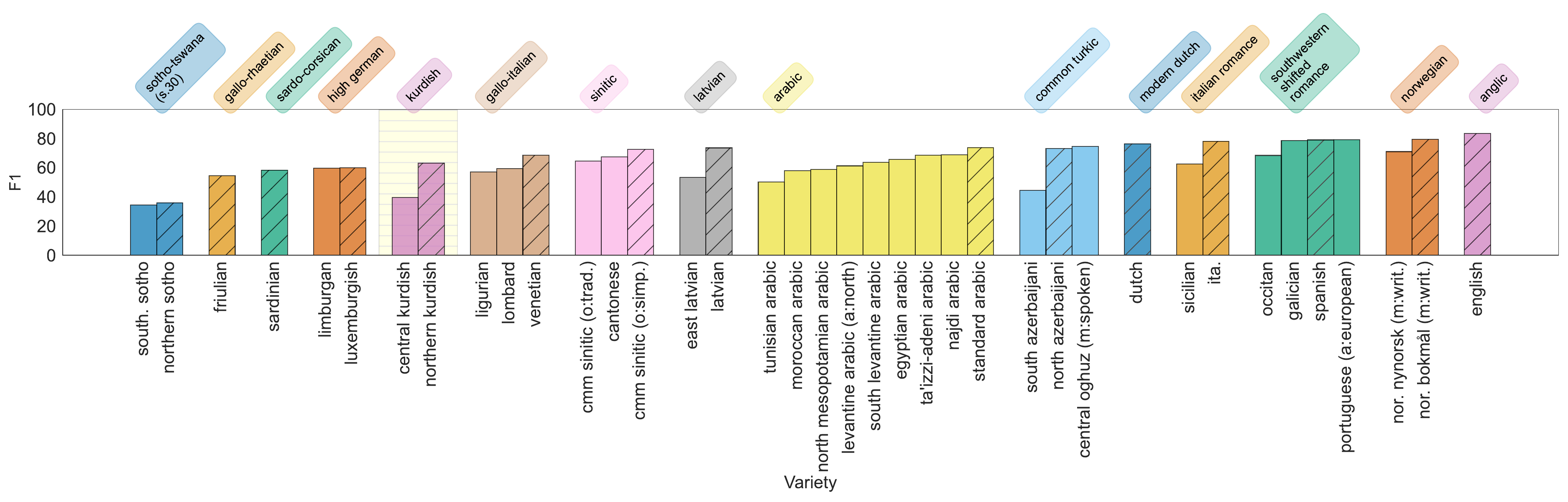}
    \caption{Natural Language Inference} 
    \label{fig:nli}
    \end{subfigure}
    \begin{subfigure}{\linewidth}
    \centering
    \includegraphics[width=1\textwidth]{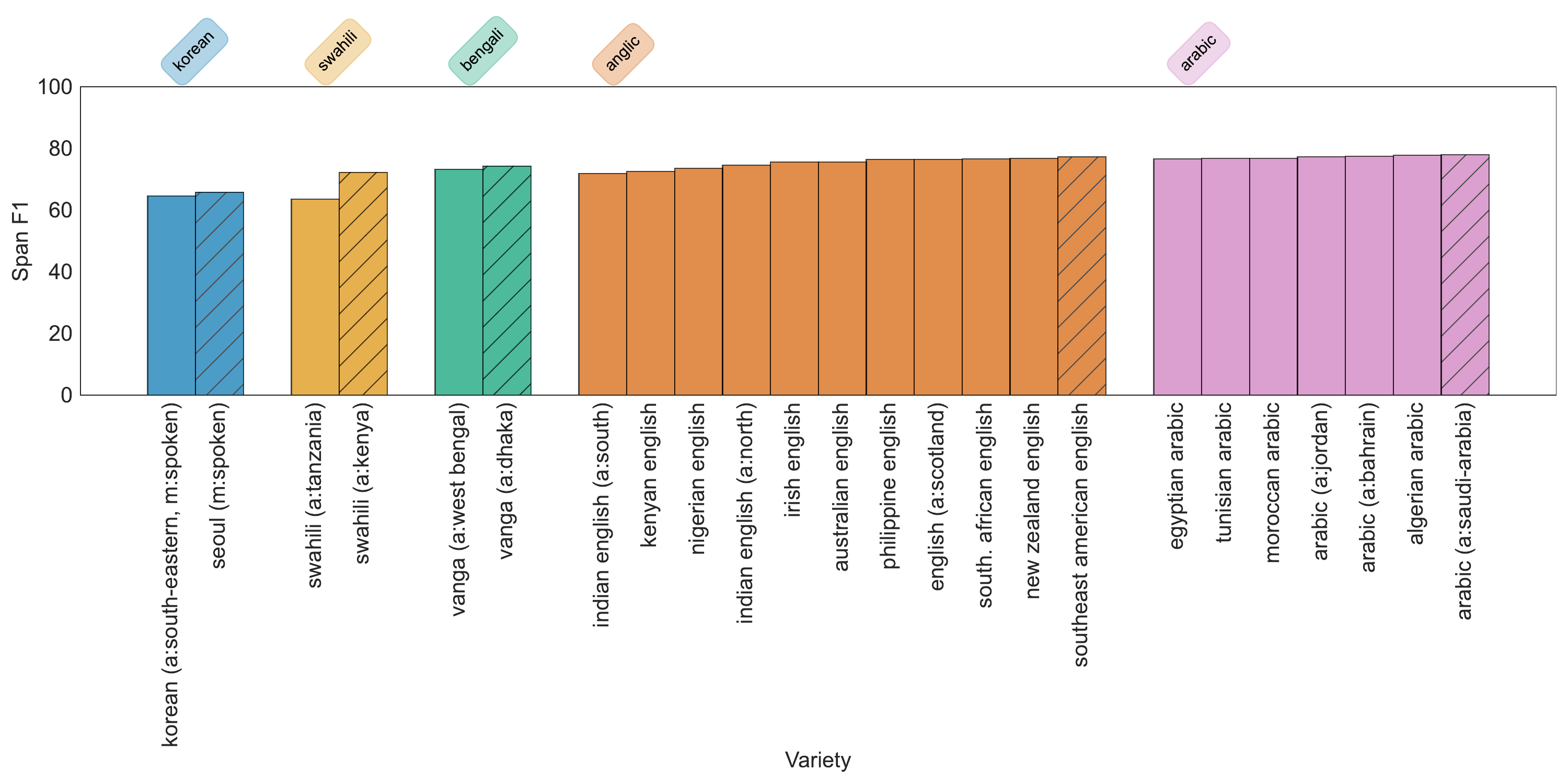}
    \caption{Extractive Question Answering}
    \label{fig:qa}
    \end{subfigure}

    \begin{subfigure}{\linewidth}
    \centering
    \includegraphics[width=1\textwidth]{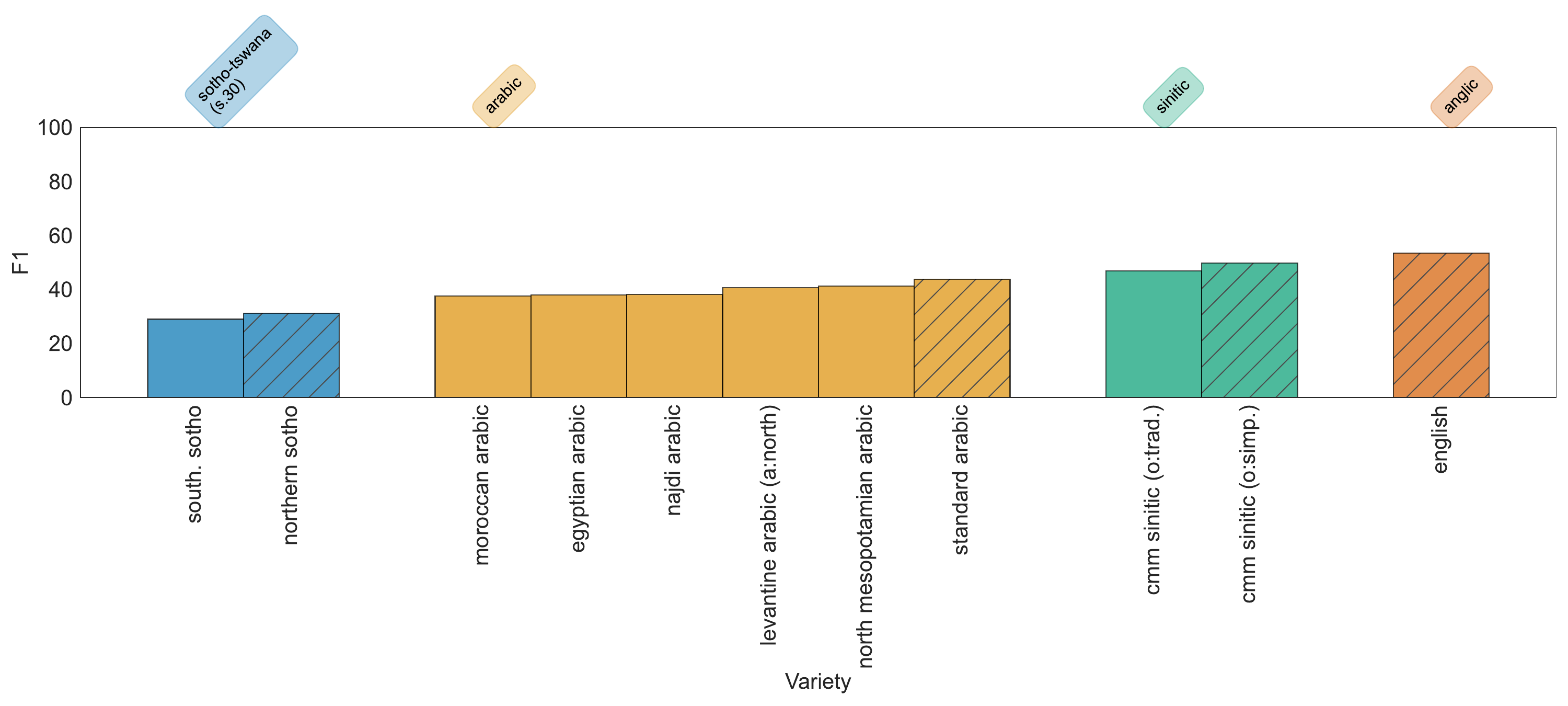}
    \caption{Multiple Choice Machine Reading Comprehension}
    \label{fig:rcmc}
    \end{subfigure}

    \caption{Task specific plot of maximum obtainable Linguistic Utility for all \dialects{}. The yellow-shaded region represents language \clusters{} having no \dialects{} seen during mBERT pertaining. The bars with colored stripes represent the standard \dialect{} of a \cluster{}. The dialect with the Rawlsian score in each cluster is that with the leftmost bar.}
    \label{fig:plot_app_2}

\end{figure*}

\begin{figure*}[ht]
    \begin{subfigure}{\linewidth}
    \centering
    \includegraphics[width=.9\textwidth]{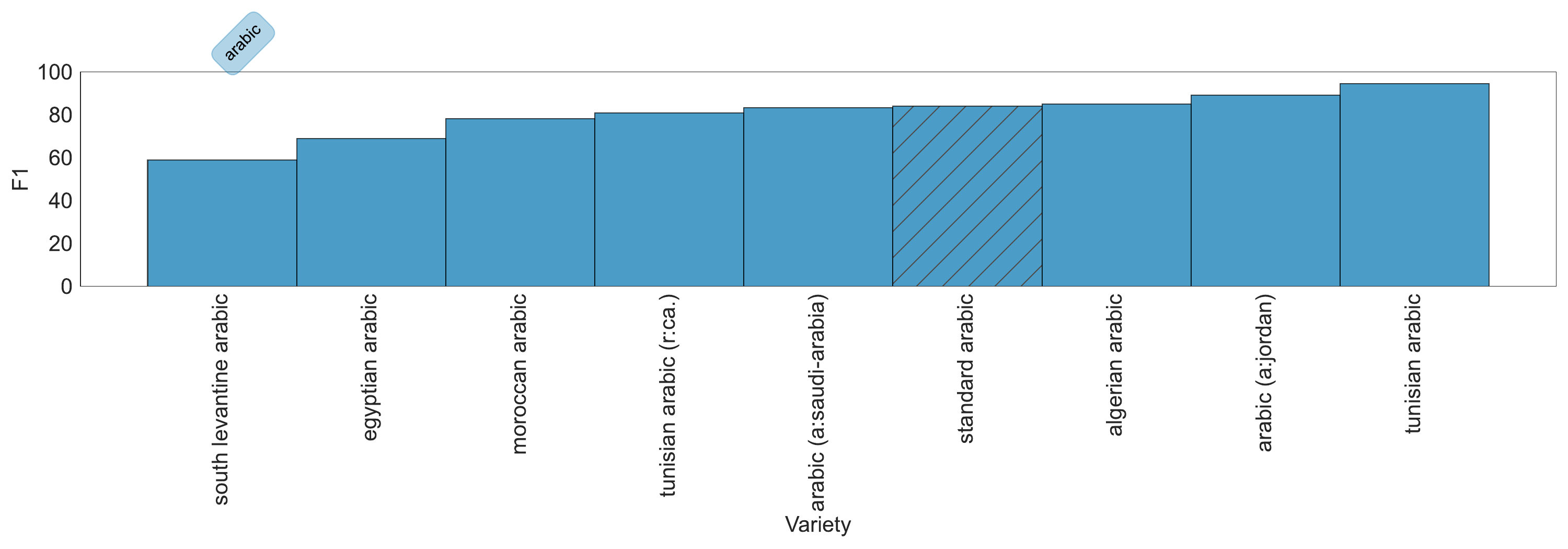}
    \caption{Sentiment Analysis} 
    \label{fig:sc}
    \end{subfigure}
    \begin{subfigure}{\linewidth}
    \centering
    \includegraphics[width=1\textwidth]{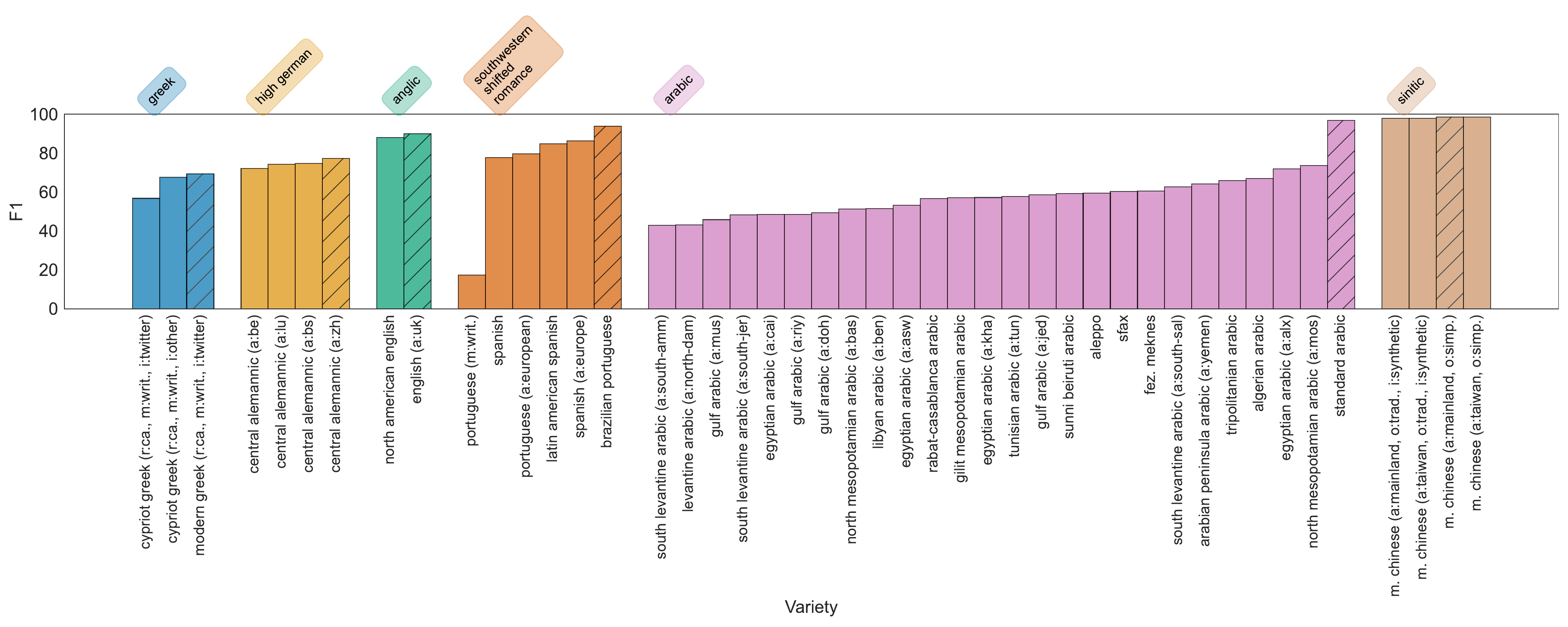}
    \caption{Dialect Identification}
    \label{fig:di}
    \end{subfigure}

    \caption{Task specific plot of maximum obtainable Linguistic Utility for all \dialects{}. The yellow-shaded region represents language \clusters{} having no \dialects{} seen during mBERT pertaining. The bars with colored stripes represent the standard \dialect{} of a \cluster{}. The dialect with the Rawlsian score in each cluster is that with the leftmost bar.}
    \label{fig:plot_app_3}

\end{figure*}

\begin{figure*}[ht]
    \begin{subfigure}{\linewidth}
    \centering
    \includegraphics[width=1\textwidth]{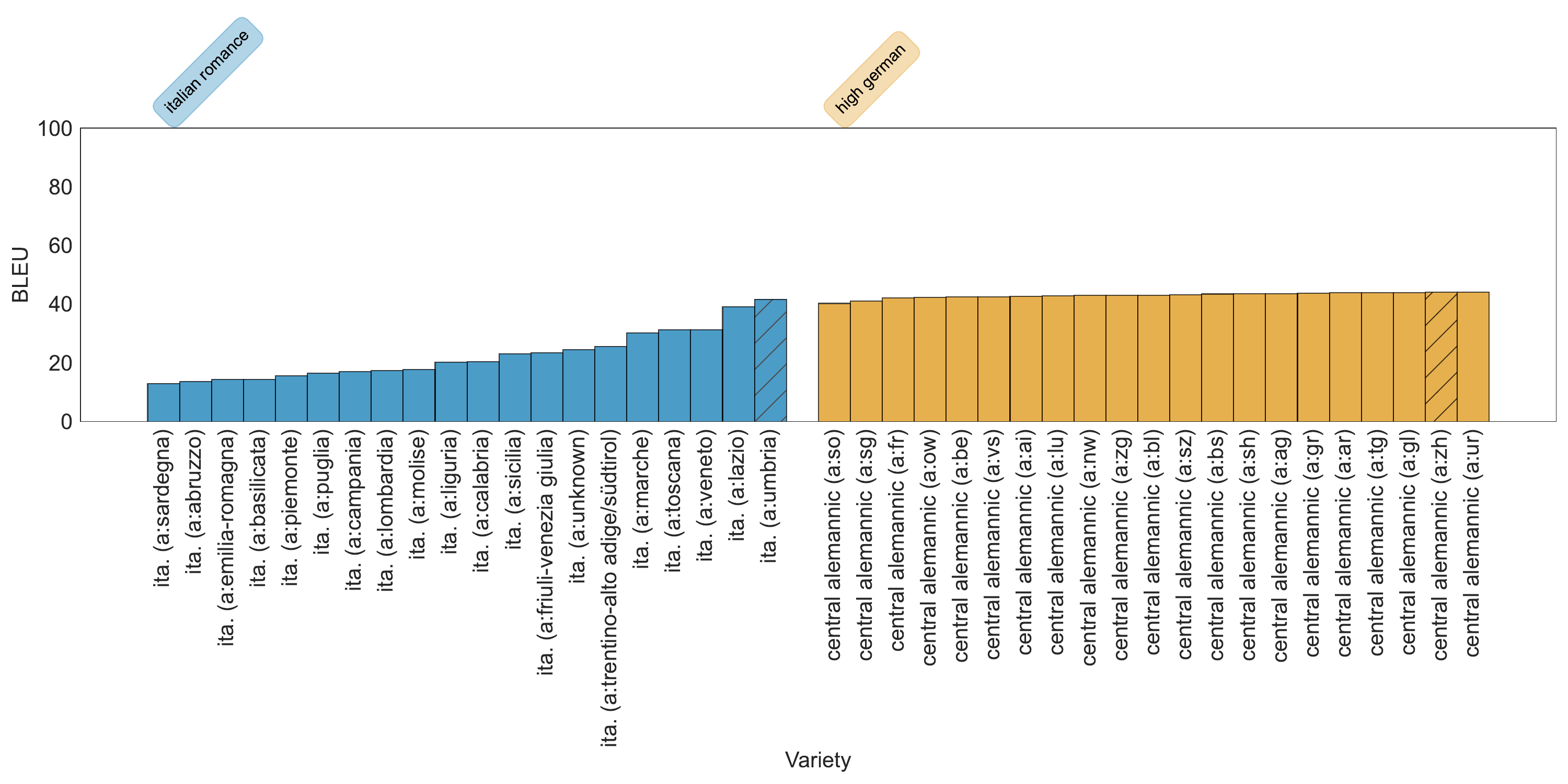}
    \caption{Machine Translation (Region level aggregation)}
    \label{fig:mt-region}
    \end{subfigure}
    
    \begin{subfigure}{\linewidth}
    \centering
    \includegraphics[width=1\textwidth]{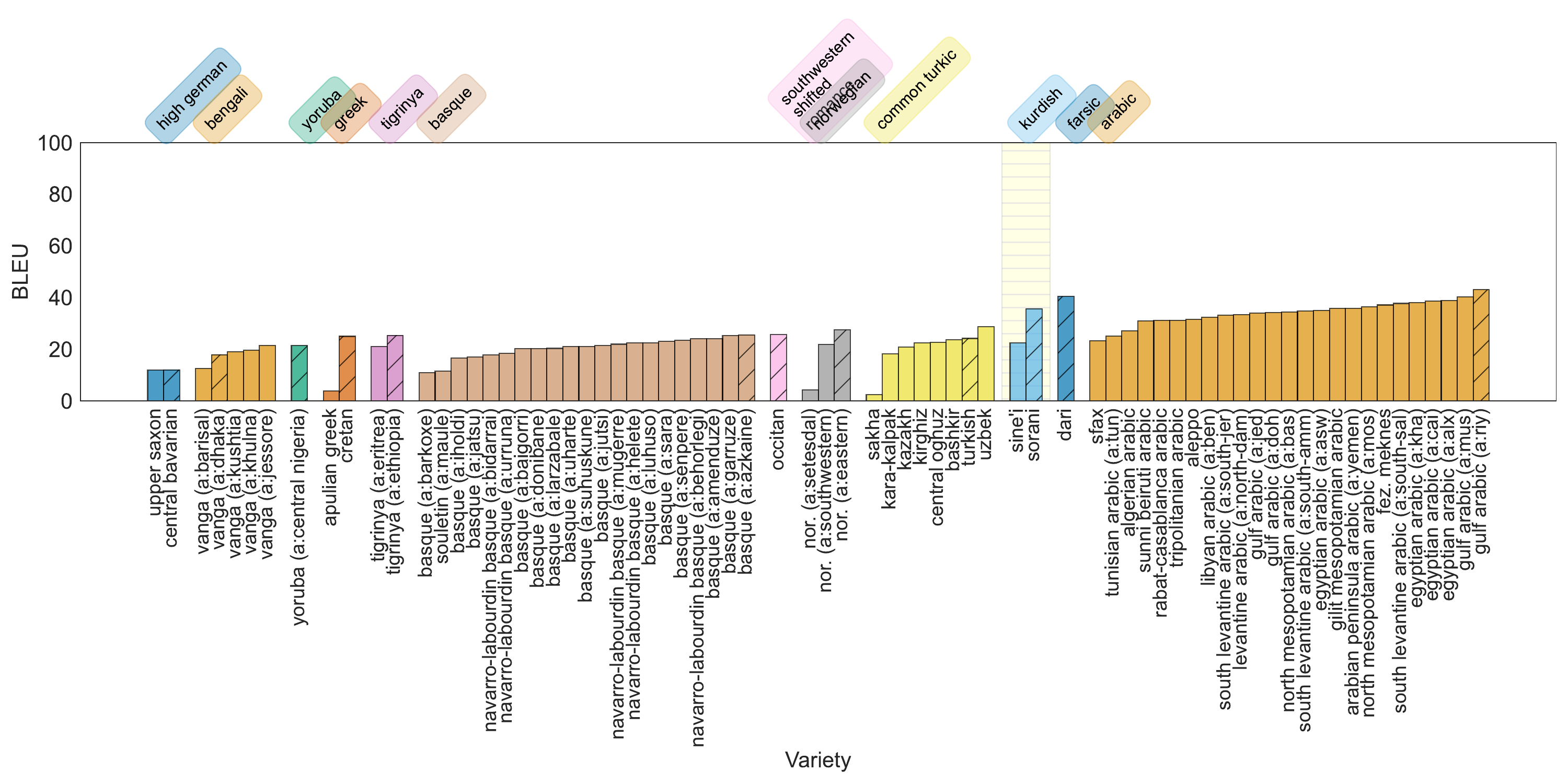}
    \caption{Machine Translation (Variety level)}
    \label{fig:mt-dialect}
    \end{subfigure}

    \caption{Task specific plot of maximum obtainable Linguistic Utility for all \dialects{}. The yellow-shaded region represents language \clusters{} having no \dialects{} seen during mBERT pertaining. The bars with colored stripes represent the standard \dialect{} of a \cluster{}. The dialect with the Rawlsian score in each cluster is that with the leftmost bar.}
    \label{fig:plot_app_4}

\end{figure*}

\subsection{Dialectal Gap visualizations (zeroshot)}
\label{app:dialect_gap_plots}
\begin{figure*}[!t]
\begin{subfigure}[b]{0.33\textwidth}
    \centering
    \includegraphics[width=\linewidth]{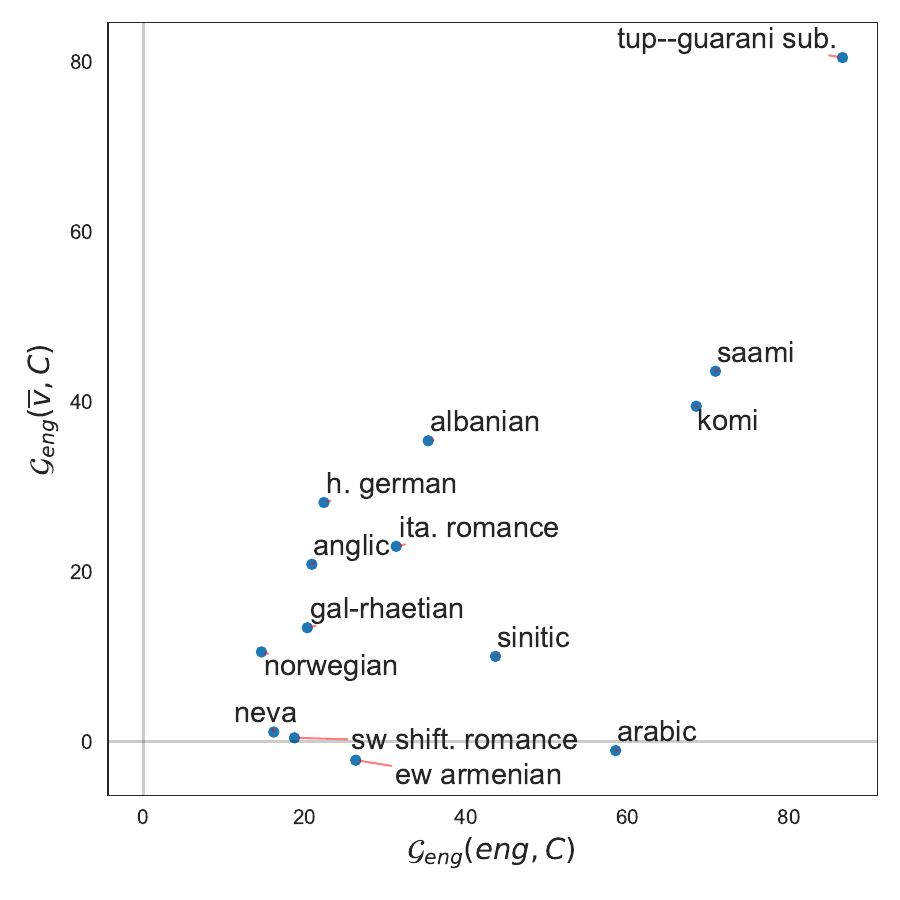}
    \caption{Parts-of-speech (POS) tagging}
    \label{fig:pos-gap}
\end{subfigure}
\begin{subfigure}[b]{0.33\textwidth}
    \centering
    \includegraphics[width=\linewidth]{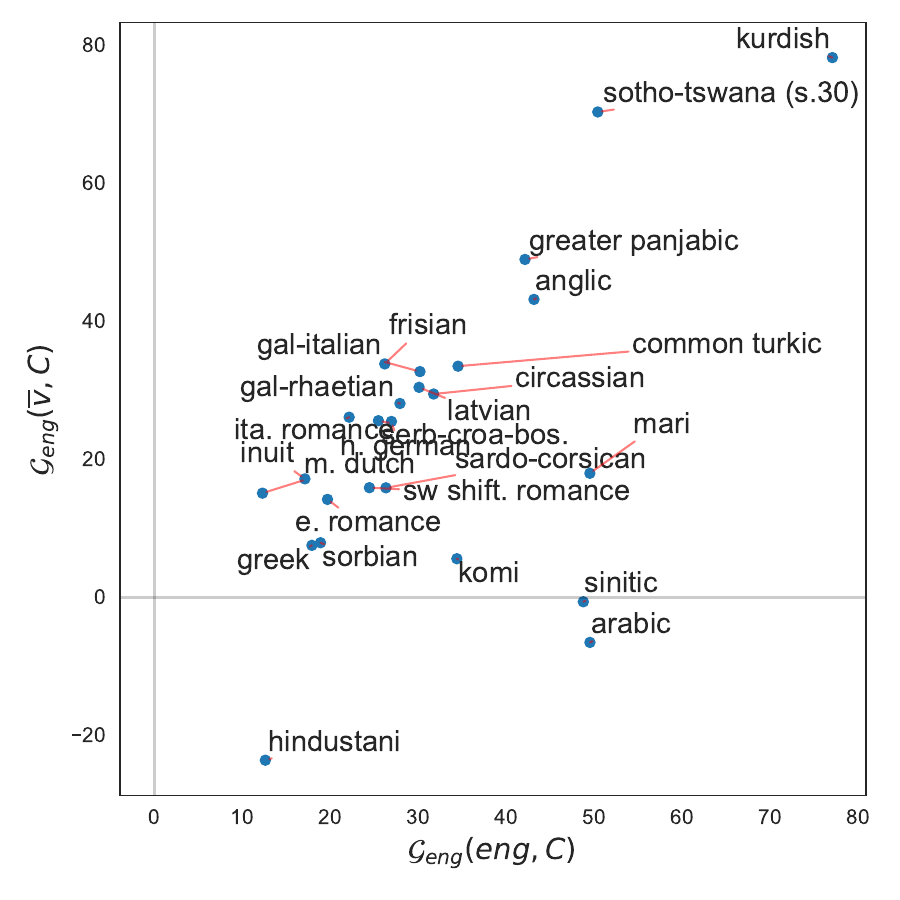}
    \caption{Named entity recognition (NER)}
    \label{fig:ner-gap}
\end{subfigure}
\begin{subfigure}[b]{0.33\textwidth}
    \centering
    \includegraphics[width=\linewidth]{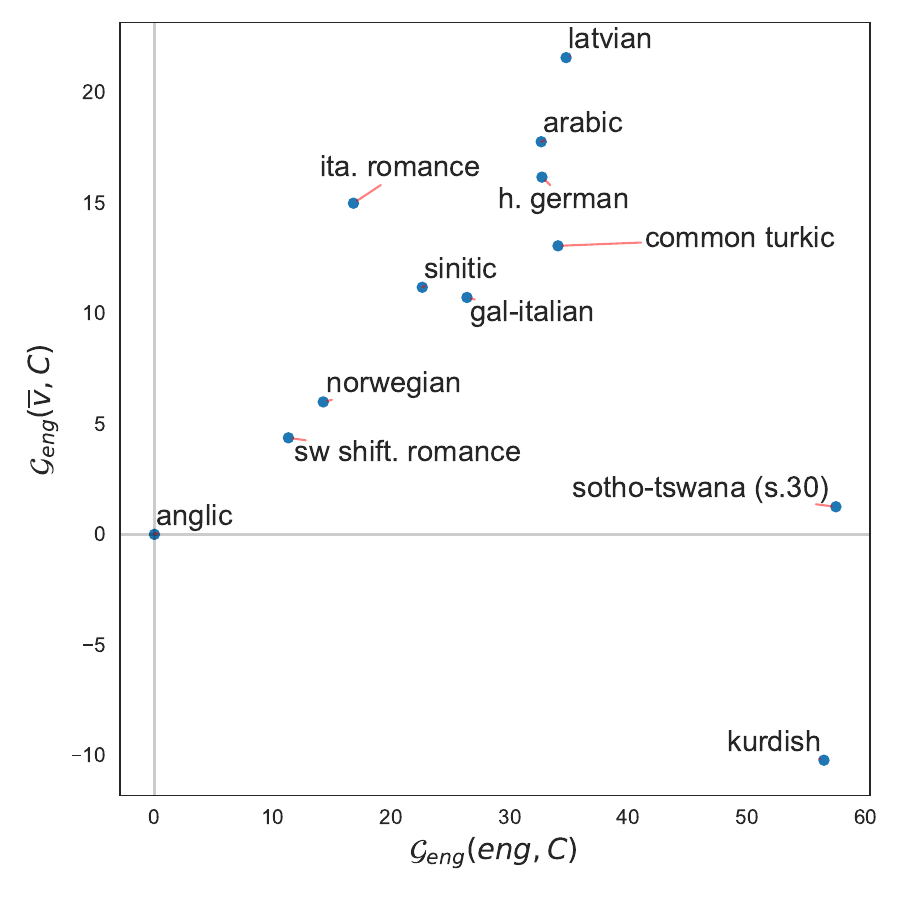}
    \caption{Natural language inference (NLI)}
    \label{fig:nli-gap}
\end{subfigure}
\caption{Dialectal Gap visualization for language \clusters{} utilizing zero-shot cross-lingual transfer from Standard English. The x-axis is for \cluster{} gap while comparing against Standard English \dialect{}. Values far from zero have a larger performance gap from English. The y-axis is for aggregated cluster gap while comparing against standard \cluster{} \dialect{} and values far from zero have larger within cluster gap. Ideally, we want both of them to be close to zero.}
\label{fig:gap-plots_app}
\end{figure*}


\section{Evaluation results}
\label{app:detailed_result_tables}
\begin{table*}
\scriptsize
\centering
\begin{tabular}{lp{3cm}p{3cm}p{1cm}p{1.2cm}p{1cm}p{1.2cm}}
\toprule
\cluster{} & \dialect{} &UD-code &               Zeroshot (mBERT) &   Zeroshot (XLM-R) &  Finetune (mBERT) & Finetune (XLM-R)\\

\midrule
\multirow{2}{*}{albanian} & albanian &                    UD\_Albanian-TSA &              81.78 &              83.08 &                  - &                  - \\
                & gheg albanian &                        UD\_Gheg-GPS &              38.14 &              43.50 &                  - &                  - \\
\cmidrule{1-7}
\multirow{3}{*}{anglic} & english &                     UD\_English-EWT &              91.55 &              91.10 &              91.55 &               91.1 \\
                & singlish &                           singlish &              69.88 &              68.55 &              82.36 &               11.0 \\
                & african american vernacular english &                         TwitterAAE &              50.53 &              52.53 &                  - &                  - \\
\cmidrule{1-7}
\multirow{3}{*}{arabic} & standard arabic &                     UD\_Arabic-PADT &              50.87 &              55.91 &              88.47 &               2.48 \\
                & south levantine arabic &    UD\_South\_Levantine\_Arabic-MADAR &              49.94 &              45.75 &                  - &                  - \\
\bottomrule
\end{tabular}
\caption{Dependency parsing evaluation report comprising zeroshot score and in-language finetuning. We report UAS as evaluation score. 
Zeroshot scores are evaluated using model finetuned on Standerd English. If training data is not available, we skip those 
languages (mentioned as '-') for in-language finetuning.}
\label{tab:dep_full}
\end{table*}

\begin{table*}[ht]
\scriptsize
\centering
\begin{tabular}{lp{3cm}llp{1cm}p{1cm}p{1cm}p{1cm}}
\toprule
  \cluster{}   &              \dialect{}        &                     Dataset-code & dataset &  Zeroshot (mBERT) &   Zeroshot (XLM-R) &  Finetune (mBERT) & Finetune (XLM-R)\\
 &  &                         &         &                   &                   &              &              \\
 \midrule
\multirow{2}{*}{albanian} & albanian &                    UD\_Albanian-TSA &      ud &             75.80 &             84.41 &            - &            - \\
                & gheg albanian &                        UD\_Gheg-GPS &      ud &             48.96 &             55.84 &            - &            - \\
\cmidrule{1-8}
\multirow{2}{*}{anglic} & english &                     UD\_English-EWT &      ud &             96.41 &             97.16 &        96.41 &        97.16 \\
                & singlish &                           singlish &      ud &             76.27 &             77.55 &        87.38 &        87.96 \\
\cmidrule{1-8}
\multirow{6}{*}{arabic} & south levantine arabic &    UD\_South\_Levantine\_Arabic-MADAR &      ud &             51.99 &             61.84 &            - &            - \\
                & standard arabic &                     UD\_Arabic-PADT &      ud &             39.74 &             56.67 &        95.72 &        96.11 \\
\bottomrule
\end{tabular}
\caption{Parts-of-speech evaluation report comprising zeroshot score and in-language finetuning. We report F1 as evaluation score. 
Zeroshot scores are evaluated using model finetuned on Standerd English. If training data is not available, we skip those 
languages (mentioned as `-') for in-language finetuning.}
\label{tab:pos_full}
\end{table*}

\onecolumn
\scriptsize
\caption{Natural language inference (NLI) evaluation report using zeroshot cross-lingual transfer from Standard English. We report F1 as evaluation score. 
NLI uses F1 as evaluation matric. We prepare a translate-train dataset to perform this evaluation.}
\label{tab:nli}
\end{table}

\begin{table}
\centering
\scriptsize
\begin{tabular}{lllrrrrr}
\toprule
   \cluster{}     &   \dialect{}        &             target-code &  count &  Finetune (mBERT) & Finetune (XLM-R) &  Zeroshot (mBERT) &   Zeroshot (XLM-R) \\
 & &                    &        &               &               &               &               \\
\midrule
\multirow{11}{*}{anglic} & irish english &       english--irl &    494 &         73.00 &         67.98 &         68.62 &         62.45 \\
        & southeast american english &       english--usa &    494 &         73.51 &         67.95 &         68.56 &         62.97 \\
        & new zealand english &       english--nzl &    494 &         73.62 &         68.50 &         68.21 &         63.05 \\
        & southern african english &       english--zaf &    494 &         73.50 &         68.06 &         68.13 &         63.19 \\
        & english (a:scotland) &       english--gbr &    494 &         73.30 &         67.57 &         67.90 &         62.49 \\
        & australian english &       english--aus &    494 &         73.23 &         68.04 &         67.71 &         62.33 \\
\bottomrule
\end{tabular}
\caption{Extractive dialectal question answering evaluation on SD-QA development set. We report span F1 as evaluation score. 
Zeroshot scores are evaluated using model finetuned on Standerd English whereas, we use combined training set for supervised finetuning}
\label{tab:sdqa_dev}
\end{table}

\begin{table}
\centering
\scriptsize
\begin{tabular}{lp{2.5cm}lrp{1cm}p{1cm}p{1cm}p{1cm}r}
\toprule
     \cluster{}   &   \dialect{}     &             target-code &  count &  Finetune (mBERT) & Finetune (XLM-R) &  Zeroshot (mBERT) &   Zeroshot (XLM-R) &  ICL Mistral \\
 &  &                    &        &               &               &               &               &            \\
\midrule
\multirow{11}{*}{anglic} & english (a:scotland) &       english--gbr &    440 &         76.38 &         70.34 &         71.82 &         63.15 &      70.18 \\
        & southern african english &       english--zaf &    440 &         76.66 &         71.18 &         71.49 &         63.87 &      71.14 \\
        & new zealand english &       english--nzl &    440 &         76.71 &         71.39 &         71.22 &         63.69 &      70.95 \\
        & australian english &       english--aus &    440 &         75.66 &         70.89 &         71.20 &         62.28 &      69.23 \\
        & southeast american english &       english--usa &    440 &         77.26 &         71.50 &         71.17 &         63.71 &      71.76 \\
        & irish english &       english--irl &    440 &         75.52 &         70.73 &         70.92 &         62.15 &      70.64 \\
\bottomrule
\end{tabular}
\caption{Extractive dialectal question answering evaluation on SD-QA test set. We report span F1 as evaluation score. 
Zeroshot scores are evaluated using model finetuned on Standerd English whereas, we use combined training set for supervised finetuning}
\label{tab:sdqa_test}
\end{table}

\begin{table}
\centering
\scriptsize
\begin{tabular}{lp{2.5cm}p{1.5cm}rp{1cm}p{1cm}p{1cm}p{1cm}p{1cm}p{.7cm}}
\toprule
     \cluster{}         &      \dialect{}     &                 dialect-code &  support &  precision (mBERT) &  precision (XLM-R) &  recall (mBERT) &  recall (XLM-R) &  F1 (mBERT) &  F1 (XLM-R) \\
 &  &                         &          &                    &                    &                 &                 &             &             \\
\midrule
\multirow{2}{*}{anglic} & english (a:uk) &           english:en-gb &      249 &              98.10 &              89.57 &           83.13 &           71.59 &       90.00 &       79.58 \\
                             & north american english &           english:en-us &      349 &              93.27 &              88.14 &           83.38 &           82.09 &       88.05 &       85.01 \\
\cmidrule{1-10}
\multirow{26}{*}{arabic} & aleppo &              arabic:ale &      200 &              59.50 &              58.50 &           59.50 &           48.35 &       59.50 &       52.94 \\
                             & algerian arabic &              arabic:alg &      272 &              79.00 &              65.50 &           58.09 &           62.68 &       66.95 &       64.06 \\
                             & arabian peninsula arabic (a:yemen) &              arabic:san &      177 &              60.50 &              55.50 &           68.36 &           56.63 &       64.19 &       56.06 \\
                             & egyptian arabic (a:alx) &              arabic:alx &      192 &              70.50 &              74.50 &           73.44 &           66.82 &       71.94 &       70.45 \\
\bottomrule
\end{tabular}
\caption{Dialect Identification evaluation using language cluster specific datasets. We finetune a classificaiton model using either mBERT or XLM-R and then evaluate on the test data.}
\label{tab:id_all}
\end{table}

\begin{table}
\centering
\scriptsize

\begin{tabular}{lp{3cm}llrrrr}
\toprule
    \cluster{}  &     \dialect{}           &    target-code &       src &  mBERT (acc) &  XLM-R (acc) &  mBERT (F1) &  XLM-R (F1) \\
 &  &           &           &            &            &          &          \\
\midrule
\multirow{9}{*}{arabic} & standard arabic &  arb\_Arab &  arb\_Arab &      85.25 &      83.96 &    86.71 &    82.27 \\
                             & ta'izzi-adeni arabic &  acq\_Arab &  arb\_Arab &      84.96 &      82.05 &    86.44 &    81.98 \\
                             & najdi arabic &  ars\_Arab &  arb\_Arab &      84.80 &      84.39 &    87.41 &    83.33 \\
                             & north mesopotamian arabic &  acm\_Arab &  arb\_Arab &      82.97 &      80.95 &    84.77 &    80.36 \\
                             & south levantine arabic &  ajp\_Arab &  arb\_Arab &      81.82 &      80.16 &    84.16 &    79.05 \\
                             & levantine arabic (a:north) &  apc\_Arab &  arb\_Arab &      81.59 &      80.15 &    83.76 &    79.88 \\
\bottomrule
\end{tabular}
\caption{Topic classification evaluation using SIB-200 language data with dialectal presence. We report span F1 as evaluation score. 
Zeroshot scores are evaluated using model finetuned on Standerd English whereas, we use in-group training for supervised finetuning}
\label{tab:topic}
\end{table}









\begin{table}[]
    \centering
    \small
\begin{tabular}{lllrrrrr}
\toprule
       &                        &   dialect &  MBERT\_Acc &  MBERT\_F1 &  XLMR\_Acc &  XLMR\_F1 &  mistral7b \\
lang-group & variety &           &            &           &           &          &            \\
\midrule
\multirow{9}{*}{arabic} & tunisian arabic &  aeb\_arab &      94.55 &     94.56 &     94.61 &    94.62 &       73.3 \\
       & algerian arabic &  arq\_arab &      84.98 &     85.00 &     84.70 &    84.69 &       76.0 \\
       & arabic (a:jordan) &  jor\_arab &      82.96 &     82.90 &     89.07 &    89.00 &       82.2 \\
       & arabic (a:saudi-arabia) &  sau\_arab &      81.38 &     65.97 &     83.40 &    67.66 &       79.8 \\
       & tunisian arabic (r:casual) &  aeb\_latn &      80.95 &     65.65 &     79.80 &    63.70 &       62.3 \\
       & standard arabic &  arb\_arab &      80.63 &     70.01 &     83.96 &    72.91 &       65.7 \\
       & moroccan arabic &  ary\_arab &      78.08 &     61.50 &     77.41 &    55.55 &       58.4 \\
       & egyptian arabic &  arz\_arab &      67.03 &     40.00 &     69.03 &    47.89 &       50.0 \\
       & south levantine arabic &     ar\_lb &      58.38 &     34.63 &     58.90 &    32.72 &        0.0 \\
\bottomrule
\end{tabular}
    \caption{Sentiment Analysis results. In addition to, using mBERT and XLM-R as the base models, we also perform in-context learning to evaluate the performance of large language models (Mistral-7B).}
    \label{tab:sc}
\end{table}

\begin{table}
\centering
\scriptsize
\begin{tabular}{lllrrrr}
\toprule
       &                 &   dialect &  acc (mBERT) &  acc (XLM-R) &  mBERT (F1) &  XLM-R (F1) \\
lang-group & language &           &                &                &          &          \\
\midrule
anglic & english &  eng\_Latn &          52.22 &          53.56 &    51.97 &    53.44 \\
\multirow{6}{*}{arabic} & standard arabic &  arb\_Arab &          39.00 &          43.78 &    39.01 &    43.78 \\
                    & levantine arabic (a:north) &  apc\_Arab &          38.89 &          40.78 &    38.64 &    40.71 \\
                    & north mesopotamian arabic &  acm\_Arab &          38.11 &          41.33 &    37.99 &    41.35 \\
                    & moroccan arabic &  ary\_Arab &          37.00 &          37.67 &    36.94 &    37.61 \\
                    & egyptian arabic &  arz\_Arab &          36.22 &          38.00 &    36.21 &    37.98 \\
                    & najdi arabic &  ars\_Arab &          36.00 &          38.11 &    36.05 &    38.16 \\
\cmidrule{1-7}
\multirow{2}{*}{sinitic} & classical-middle-modern sinitic (o:simplified) &  zho\_Hans &          50.11 &          47.22 &    49.79 &    47.10 \\
                    & classical-middle-modern sinitic (o:traditional) &  zho\_Hant &          47.00 &          45.11 &    46.88 &    44.76 \\
\cmidrule{1-7}
\multirow{2}{*}{sotho-tswana (s.30)} & northern sotho &  nso\_Latn &          31.11 &          29.78 &    31.18 &    29.72 \\
                    & southern sotho &  sot\_Latn &          28.56 &          29.11 &    28.52 &    29.00 \\
\bottomrule
\end{tabular}
\caption{Multiple-choice machine reading comprehension evaluation using Belebele dataset languages with dialectal presence. We report span F1 as evaluation score. 
We use combined finetuning using the aggregated training data provided with Belebele evaluationd data.}
\label{tab:mrc}
\end{table}

\small
\begin{longtable}{lrrr}

\caption{Zero-shot results for Machine Translation. We evaluate NLLB\_600m and NLLB\_1\_3bn by translating each dialectal variety to English. For all languages without reference translations in English, we evaluate dialectal translations using the standard language's translation as the reference. For varieties with parallel data like Yoruba, Turkish, Farsi, Tigrinya and Bengali we use the English reference provided in the dataset.  }
\label{zero-shot-mt}\\
\toprule
language\_group &                     \dialect &  NLLB\_600m-bleu &  NLLB\_1\_3bn-bleu \\
\midrule
\endfirsthead
\caption[]{Machine Translation} \\
\toprule
language\_group &                     \dialect &  NLLB\_600m-bleu &  NLLB\_1\_3bn-bleu \\
\midrule
\endhead
\midrule
\multicolumn{4}{r}{{Continued on next page}} \\
\midrule
\endfoot

\bottomrule
\endlastfoot
\multirow{25}{*}{arabic} & gulf arabic (a:riy) &           43.07 &            43.07 \\
       & gulf arabic (a:mus) &           40.32 &            40.32 \\
       & egyptian arabic (a:alx) &           38.90 &            38.90 \\
       & egyptian arabic (a:cai) &           38.71 &            38.71 \\
       & egyptian arabic (a:kha) &           38.10 &            38.10 \\
       & south levantine arabic (a:south-sal) &           37.79 &            37.79 \\
       & fez. meknes &           37.19 &            37.19 \\
       & north mesopotamian arabic (a:mos) &           36.47 &            36.47 \\
\end{longtable}

\section{Highest performing and lowest performing varieties}
\label{app:high_low}
\begin{table*}[t]
\centering
\scriptsize
    \begin{tabular}{l|p{3cm}p{3cm}|p{3cm}p{3cm}}
\toprule
    &               \multicolumn{2}{c}{\Dialects{} with Highest Performance}                                   &                     \multicolumn{2}{c}{\Dialects{} with Lowest Performance}                           \\
Task (Dataset) & &                                  &                           &                              \\
\midrule
\multirow{5}{*}{DEP parsing (UD)} & anglic/english* &           italian romance/italian (r:formal, m:written, i:essay)* &               tupi-guarani subgroup i.a/old guarani* &                                        arabic/north african arabic \\
                 & albanian/albanian* &              southwestern shifted romance/portuguese (a:european) &                        komi/komi-zyrian (m:written)* &                       italian romance/continental southern italian \\
                 & gallo-rhaetian/french* &                          norwegian/norwegian nynorsk (m:written)* &                                   saami/skolt saami* &                               komi/komi-zyrian (m:spoken)$\dagger$ \\
                 & norwegian/norwegian bokmål (m:written)* &                  southwestern shifted romance/portuguese (i:mix)* &  tupi-guarani subgroup i.a/mbyá guaraní (a:paraguay) &                                         komi/komi-permyak$\dagger$ \\
                 & italian romance/italian* &                southwestern shifted romance/brazilian portuguese* &   tupi-guarani subgroup i.a/mbyá guaraní (a:brazil)* &                                        saami/north saami*$\dagger$ \\
\cmidrule{1-5}
\multirow{5}{*}{EQA (SD-QA-test)} & anglic/english (a:scotland) &                                              anglic/irish english &                           swahili/swahili (a:kenya)* &                                    arabic/algerian arabic$\dagger$ \\
\bottomrule
\end{tabular}
 \caption{\Dialects{} with the highest and lowest performance (in terms of raw evaluation score) on various tasks in the zero-shot setting. On the left are the top 10. We find that most of these \dialects{} are high-resource standard varieties, and only a few high-resource non-standard \dialects. At right are the bottom 10, which are mainly low-resource, non-standard \dialects{}. We use the following notations for \dialect{} type and writing system quantification: * marks standard \dialects{} (for some \clusters{}, here we consider multiple \dialects{} as standard because of the substantial resource presence of both of the \dialects{}; e.g. portuguese/european and portuguese/mix) and $\dagger$ notes mix and non-Latin writing system.}
    \label{tab:high-lang}
\end{table*}

\section{Low-resource \dialect{} performing better in zero-shot NER}
\label{app:non_std_zeroshot}
\begin{table}
\scriptsize
    \centering
    \begin{tabular}{@{}l|ll|ll@{}}
    \toprule
    \Cluster & High-resource ~\dialect & Script & Low-resource~\dialect & Script \\
    \midrule
         Hindustani & Hindi&Devanagari&Fiji hindi&Latin \\
         Greater panjabic & Eastern panjabi & Devanagari & Western panjabi & Arabic \\
         Sotho-tswana (s.30) & Northern sotho &Latin&Southern sotho& Latin\\
         \bottomrule
    \end{tabular}
    \caption{NER cases where low-resource varieties perform better in zero-shot. The script plays a significant role here as most of the cases, high-resource scripts such as Latin, and Arabic could uplift the performance of a low-resource \dialect{} over a high-resource non-Latin script.}
    \label{tab:script}
\end{table}

\section{\Cluster{} level Result Summaries with \textit{Demographic Utility} and Standard Deviation report}
\label{app:c_summary}
\begin{table}
\centering \small
\caption{Language \clusters{} with their \lingu{}, demographic utility, \dialect{} with minimum \lingu{} and standard deviation. Task: DEP Parsing}
\label{tab:summery-DEP. Parsing}
\begin{tabular}{lp{1cm}p{1.5cm}lr}
\toprule
\clusters{} &  \lingu{} & \demu{} & \dialect{} (minimum ling. u.) &  standard deviation \\
\midrule
                    albanian &      63.3 &                69.1 &                        (gheg albanian, 43.5) &               28.0 \\
                      anglic &      75.5 &                90.7 &  (african american vernacular english, 52.5) &               20.4 \\
                      arabic &      64.4 &                83.3 &               (south levantine arabic, 49.9) &               21.0 \\
    eastern-western armenian &      88.2 &                87.5 &                     (eastern armenian, 87.0) &                1.6 \\
               gallo-italian &      50.2 &                50.2 &                             (ligurian, 50.2) &                  - \\
              gallo-rhaetian &      87.7 &                93.4 &                     (french (a:paris), 77.5) &                8.8 \\
                 high german &      56.6 &                76.5 &             (central alemannic (a:zh), 36.8) &               28.1 \\
             italian romance &      75.4 &                76.5 &         (continental southern italian, 50.0) &               17.3 \\
                        komi &      30.2 &                30.1 &              (komi-zyrian (m:written), 27.6) &                2.4 \\
                   norwegian &      88.4 &                   - & (norwegian nynorsk (m:written, i:old), 78.4) &                8.6 \\
                       saami &      47.9 &                67.0 &                          (skolt saami, 28.3) &               27.8 \\
                    sabellic &      33.2 &                   - &                              (umbrian, 33.2) &                  - \\
                     sinitic &      65.4 &                   - &                    (classical chinese, 46.7) &               20.8 \\
southwestern shifted romance &      87.2 &                91.6 &              (portuguese (a:european), 77.3) &                7.9 \\
   tupi-guarani subgroup i.a &      17.2 &                22.0 &               (mbyá guaraní (a:brazil), 9.0) &               10.5 \\
             west low german &      40.8 &                40.8 &                      (west low german, 40.8) &                  - \\
\bottomrule
\end{tabular}
\end{table}

\begin{table}
\centering \small
\caption{Language \clusters{} with their \lingu{}, demographic utility, \dialect{} with minimum \lingu{} and standard deviation. Task: POS Tagging}
\label{tab:summery-POS Tagging}
\begin{tabular}{p{3cm}rlp{4cm}r}
\toprule
\clusters{} &  \lingu{} & \demu{} & \dialect{} (minimum ling. u.) &  standard deviation \\
\midrule
   albanian &      70.1 &                74.3 &                                               (gheg albanian, 55.8) &               20.2 \\
                      anglic &      92.6 &                97.1 &                                                    (singlish, 88.0) &                6.5 \\
                      arabic &      61.9 &                81.4 &                                            (levantine arabic, 43.8) &               18.9 \\
    eastern-western armenian &      93.8 &                93.0 &                                            (eastern armenian, 92.4) &                2.0 \\
               gallo-italian &      58.9 &                58.9 &                                                    (ligurian, 58.9) &                  - \\
              gallo-rhaetian &      96.4 &                96.9 &                                   (old french (842-ca. 1400), 95.6) &                0.7 \\
                 high german &      75.5 &                88.4 &                                    (central alemannic (a:zh), 62.6) &               18.2 \\
             italian romance &      82.6 &                80.9 &                                (continental southern italian, 57.1) &               17.9 \\
                        komi &      41.8 &                41.6 &                                     (komi-zyrian (m:written), 35.1) &                6.0 \\
                        neva &      75.4 &                97.7 &                                   (neva (a:south-west trans), 61.7) &               13.9 \\
                   norwegian &      97.6 &                   - &                        (norwegian nynorsk (m:written, i:old), 95.7) &                1.6 \\
                       saami &      56.5 &                78.3 &                                                 (skolt saami, 34.1) &               31.6 \\
                    sabellic &      11.9 &                   - &                                                     (umbrian, 11.9) &                  - \\
                     sinitic &      84.5 &                   - & (classical-middle-modern sinitic (a:hongkong, o:traditional), 69.0) &               13.7 \\
southwestern shifted romance &      90.5 &                95.7 &                                                     (occitan, 76.8) &               10.5 \\
   tupi-guarani subgroup i.a &      13.7 &                18.5 &                                      (mbyá guaraní (a:brazil), 1.9) &               13.7 \\
             west low german &      69.7 &                69.7 &                                             (west low german, 69.7) &                  - \\
\bottomrule
\end{tabular}
\end{table}

\begin{table}
\centering \small
\caption{Language \clusters{} with their \lingu{}, demographic utility, \dialect{} with minimum \lingu{} and standard deviation. Task: NER}
\label{tab:summery-NER}
\begin{tabular}{p{2.5cm}rllr}
\toprule
\clusters{} &  \lingu{} & \demu{} & \dialect{} (minimum ling. u.) &  standard deviation \\
\midrule
     anglic &      57.3 &                83.9 &        (jamaican creole english, 0.0) &               40.9 \\
                      arabic &      81.2 &                86.0 &               (egyptian arabic, 73.3) &               11.1 \\
                  circassian &      57.4 &                54.1 &                     (kabardian, 47.5) &               14.0 \\
               common turkic &      67.9 &                85.1 &             (south azerbaijani, 31.7) &               28.8 \\
             eastern romance &      75.9 &                79.3 &                     (moldavian, 64.6) &               16.0 \\
                     frisian &      66.8 &                80.4 &              (northern frisian, 54.8) &               13.3 \\
               gallo-italian &      68.6 &                68.0 &            (emiliano-romagnolo, 45.5) &               17.3 \\
              gallo-rhaetian &      71.8 &                90.3 &                       (walloon, 46.3) &               16.5 \\
            greater panjabic &      54.9 &                53.2 &               (eastern panjabi, 45.2) &               13.6 \\
                       greek &      81.9 &                90.1 &                        (pontic, 72.6) &               13.2 \\
                 high german &      71.5 &                88.0 &           (pennsylvania german, 41.8) &               15.9 \\
                  hindustani &      86.6 &                88.1 &                    (fiji hindi, 85.2) &                2.1 \\
                       inuit &      69.7 &                63.5 &                   (kalaallisut, 63.2) &                9.2 \\
             italian romance &      73.0 &                88.4 &  (continental southern italian, 61.5) &               13.6 \\
                        komi &      55.2 &                56.6 &                  (komi-permyak, 53.6) &                2.2 \\
                     kurdish &      47.5 &                54.8 &               (central kurdish, 35.5) &               17.0 \\
                     latvian &      71.7 &                88.7 &                  (east latvian, 50.1) &               30.5 \\
                        mari &      52.6 &                47.4 &                  (eastern mari, 46.7) &                8.3 \\
                modern dutch &      83.8 &                91.2 &                        (zeeuws, 79.5) &                7.1 \\
                   norwegian &      89.7 &                   - & (norwegian nynorsk (m:written), 87.9) &                1.5 \\
              sardo-corsican &      75.9 &                79.8 &                      (corsican, 70.6) &                7.5 \\
    serbian-croatian-bosnian &      83.1 &                81.0 &              (serbian standard, 65.9) &               11.9 \\
                     sinitic &      64.3 &                78.3 &                 (hakka chinese, 40.0) &               16.8 \\
                     sorbian &      68.3 &                67.4 &                 (upper sorbian, 65.5) &                3.9 \\
         sotho-tswana (s.30) &      49.5 &                41.1 &                (northern sotho, 29.7) &               28.0 \\
southwestern shifted romance &      74.0 &                92.1 &                     (mirandese, 49.3) &               18.1 \\
             west low german &      80.3 &                80.3 &               (west low german, 80.3) &                  - \\
\bottomrule
\end{tabular}
\end{table}

\begin{table}
\centering \small
\caption{Language \clusters{} with their \lingu{}, demographic utility, \dialect{} with minimum \lingu{} and standard deviation. Task: NLI}
\label{tab:summery-NLI}
\begin{tabular}{p{3cm}rlp{4cm}r}
\toprule
\clusters{} &  \lingu{} & \demu{} & \dialect{} (minimum ling. u.) & standard deviation \\
\midrule
    anglic &      83.4 &                83.4 &                                         (english, 83.4) &                  - \\
                      arabic &      63.3 &                70.0 &                                 (tunisian arabic, 50.2) &                7.1 \\
               common turkic &      64.1 &                62.3 &                               (south azerbaijani, 44.6) &               16.9 \\
               gallo-italian &      61.7 &                63.6 &                                        (ligurian, 57.2) &                6.0 \\
              gallo-rhaetian &      54.6 &                54.6 &                                        (friulian, 54.6) &                  - \\
                 high german &      59.9 &                59.8 &                                       (limburgan, 59.7) &                0.2 \\
             italian romance &      70.4 &                77.1 &                                        (sicilian, 62.7) &               11.0 \\
                     kurdish &      51.4 &                55.5 &                                 (central kurdish, 39.6) &               16.7 \\
                     latvian &      63.6 &                71.5 &                                    (east latvian, 53.5) &               14.2 \\
                modern dutch &      76.5 &                76.5 &                                           (dutch, 76.5) &                  - \\
                   norwegian &      75.3 &                   - &                   (norwegian nynorsk (m:written), 71.1) &                6.0 \\
              sardo-corsican &      58.3 &                58.3 &                                       (sardinian, 58.3) &                  - \\
                     sinitic &      68.2 &                67.4 & (classical-middle-modern sinitic (o:traditional), 64.5) &                4.1 \\
         sotho-tswana (s.30) &      35.3 &                35.6 &                                  (southern sotho, 34.6) &                1.0 \\
southwestern shifted romance &      76.3 &                79.1 &                                         (occitan, 68.5) &                5.2 \\
\bottomrule
\end{tabular}
\end{table}

\begin{table}
\centering \small
\caption{Language \clusters{} with their \lingu{}, demographic utility, \dialect{} with minimum \lingu{} and standard deviation. Task: TC}
\label{tab:summery-TC}
\begin{tabular}{lrlp{4cm}r}
\toprule
\clusters{} &  \lingu{} & \demu{} & \dialect{} (minimum ling. u.) & standard deviation \\
\midrule
     anglic &      89.7 &                89.7 &                                        (english, 89.7) &                  - \\
                      arabic &      84.5 &                85.7 &                                (moroccan arabic, 79.1) &                2.4 \\
               common turkic &      78.7 &                77.3 &                              (south azerbaijani, 69.7) &                7.9 \\
               gallo-italian &      73.8 &                73.7 &                                        (lombard, 70.6) &                3.0 \\
              gallo-rhaetian &      68.8 &                68.8 &                                       (friulian, 68.8) &                  - \\
                 high german &      74.5 &                72.7 &                                      (limburgan, 71.1) &                4.8 \\
             italian romance &      81.4 &                86.8 &                                       (sicilian, 75.2) &                8.8 \\
                     kurdish &      43.8 &                52.3 &                                (central kurdish, 19.4) &               34.5 \\
                     latvian &      75.6 &                82.0 &                                   (east latvian, 67.4) &               11.5 \\
                modern dutch &      89.6 &                89.6 &                                          (dutch, 89.6) &                  - \\
                   norwegian &      86.7 &                   - &                   (norwegian bokmål (m:written), 84.1) &                3.6 \\
              sardo-corsican &      71.0 &                71.0 &                                      (sardinian, 71.0) &                  - \\
                     sinitic &      89.5 &                89.5 & (classical-middle-modern sinitic (o:simplified), 89.2) &                0.3 \\
         sotho-tswana (s.30) &      37.8 &                36.9 &                                 (northern sotho, 35.6) &                3.1 \\
southwestern shifted romance &      87.2 &                86.9 &                                        (occitan, 84.1) &                2.3 \\
\bottomrule
\end{tabular}
\end{table}

\begin{table}
\centering \small
\caption{Language \clusters{} with their \lingu{}, demographic utility, \dialect{} with minimum \lingu{} and standard deviation. Task: DId}
\label{tab:summery-DId}
\begin{tabular}{lrlp{4cm}r}
\toprule
\clusters{} &  \lingu{} & \demu{} &           \dialect{} (minimum ling. u.) &  standard deviation \\
\midrule
     anglic &      89.0 &                88.4 &                                    (north american english, 88.0) &                 1.4 \\
                      arabic &      58.1 &                89.0 &                      (south levantine arabic (a:south-amm), 43.0) &                11.3 \\
                       greek &      64.6 &                   - &            (cypriot greek (r:casual, m:written, i:twitter), 56.8) &                 6.8 \\
                 high german &      74.5 &                   - &                                  (central alemannic (a:be), 72.0) &                 2.1 \\
                     sinitic &      98.3 &                98.3 & (mandarin chinese (a:mainland, o:traditional, i:synthetic), 97.9) &                 0.4 \\
southwestern shifted romance &      73.3 &                82.7 &                                    (portuguese (m:written), 17.4) &                27.9 \\
\bottomrule
\end{tabular}
\end{table}

\begin{table}
\centering \small
\caption{Language \clusters{} with their \lingu{}, demographic utility, \dialect{} with minimum \lingu{} and standard deviation. Task: SA}
\label{tab:summery-SC}
\begin{tabular}{lrrlr}
\toprule
\clusters{} &  \lingu{} &  \demu{} & \dialect{} (minimum ling. u.) &  standard deviation \\
\midrule
     arabic &      80.3 &                 81.4 &       (levantine/south, 58.9) &                10.7 \\
\bottomrule
\end{tabular}
\end{table}

\begin{table}
\centering \small
\caption{Language \clusters{} with their \lingu{}, demographic utility, \dialect{} with minimum \lingu{} and standard deviation. Task: MRC}
\label{tab:summery-MRC}
\begin{tabular}{lrlp{4cm}r}
\toprule
\clusters{} &  \lingu{} & \demu{} & \dialect{} (minimum ling. u.) & standard deviation \\
\midrule
     anglic &      53.4 &                53.4 &                                         (english, 53.4) &                  - \\
             arabic &      39.9 &                42.1 &                                 (moroccan arabic, 37.6) &                2.4 \\
            sinitic &      48.3 &                   - & (classical-middle-modern sinitic (o:traditional), 46.9) &                2.1 \\
sotho-tswana (s.30) &      30.1 &                30.5 &                                  (southern sotho, 29.0) &                1.5 \\
\bottomrule
\end{tabular}
\end{table}

\begin{table}
\centering \small
\caption{Language \clusters{} with their \lingu{}, demographic utility, \dialect{} with minimum \lingu{} and standard deviation. Task: EQA}
\label{tab:summery-EQA}
\begin{tabular}{lrrlr}
\toprule
\clusters{} &  \lingu{} &  \demu{} & \dialect{} (minimum ling. u.) &  standard deviation \\
\midrule
     anglic &      75.2 &                75.4 &           (indian english (a:south), 71.9) &                 1.8 \\
     arabic &      77.2 &                77.0 &                    (egyptian arabic, 76.5) &                 0.6 \\
    bengali &      73.8 &                   - &              (vanga (a:west bengal), 73.3) &                 0.7 \\
     korean &      65.2 &                64.6 & (korean (a:south-eastern, m:spoken), 64.6) &                 0.8 \\
    swahili &      67.9 &                64.1 &               (swahili (a:tanzania), 63.5) &                 6.2 \\
\bottomrule
\end{tabular}
\end{table}

\section{In-Context Learning Details}\label{sec:icl_prompts}
\subsection{Prompts}

We adapt the prompts from Super-NaturalInstructions \cite{wang-etal-2022-super} for our in-context learning experiments.  

\paragraph{Sentiment Analysis.} For sentiment analysis we provide 4 few-shot examples in the prompt. The prompt template is given below:

\hfill

\noindent \texttt{In this task, you are given a piece of text. Your task is to classify the sentiment of the text based on its content.}

\hfill

\noindent \texttt{Sentence: <Sentence Example 1>}

\noindent \texttt{Label: <Label for Example 1, Positive, negative, neutral>}

\hfill

\noindent \texttt{$\cdots$}

\hfill

\noindent\texttt{Sentence: <Sentence Example $k$>}

\noindent\texttt{Label: <Label for Example $k$, Positive, negative, neutral>}

\hfill

\noindent\texttt{Sentence: <Test Example Input>}

\noindent\texttt{Label:}

\hfill

\paragraph{Extractive Question Answering.} We provide 2 few-shot examples i.e. $k = 2$ due to the long-form nature of text for this task.

\hfill

\noindent\texttt{This task is about writing a correct answer for the reading comprehension task. Based on the information provided in a given passage, you should identify the shortest continuous text span from the passage that serves as an answer to the given question. Avoid answers that are incorrect or provides incomplete justification for the question.}

\hfill

\noindent\texttt{Passage: <Passage for Example 1>}

\noindent\texttt{Question: <Question for Example 1>}

\noindent\texttt{Answer: <Answer for Example 1>}

\hfill

\noindent\texttt{$\cdots$}

\hfill

\noindent\texttt{Passage: <Passage for Example $k$>}

\noindent\texttt{Question: <Question for Example $k$>}

\noindent\texttt{Answer: <Answer for Example $k$>}

\hfill

\noindent\texttt{Passage: <Passage for Test Example>}

\noindent\texttt{Question: <Question for Test Example>}

\noindent\texttt{Answer:}

\end{document}